\documentclass{article}

\usepackage{PRIMEarxiv}

\usepackage[utf8]{inputenc} 
\usepackage[T1]{fontenc}    
\usepackage{hyperref}       
\usepackage{url}            
\usepackage{booktabs}       
\usepackage{amsfonts}       
\usepackage{nicefrac}       
\usepackage{microtype}      
\usepackage{lipsum}
\usepackage{fancyhdr}       
\usepackage{graphicx}       
\graphicspath{{media/}}     

\usepackage{amsopn}
\usepackage{amsmath}
\usepackage{cleveref}
\usepackage{amsthm}

\usepackage{epstopdf}
\usepackage{algorithmic}

\usepackage[para,online,flushleft]{threeparttable}

\ifpdf
  \DeclareGraphicsExtensions{.eps,.pdf,.png,.jpg}
\else
  \DeclareGraphicsExtensions{.eps}
\fi

\newcommand{\overbar}[1]{\mkern 1.5mu\overline{\mkern-1.5mu#1\mkern-1.5mu}\mkern 1.5mu}

\newtheorem{innercustomgeneric}{\customgenericname}
\providecommand{\customgenericname}{}
\newcommand{\newcustomtheorem}[2]{%
  \newenvironment{#1}[1]
  {%
   \renewcommand\customgenericname{#2}%
   \renewcommand\theinnercustomgeneric{##1}%
   \innercustomgeneric
  }
  {\endinnercustomgeneric}
}

\newcustomtheorem{customthm}{Theorem}
\newcustomtheorem{customlemma}{Lemma}
\newcustomtheorem{customcondition}{Condition}

\newtheorem*{theorem*}{Theorem}
\newtheorem*{proposition*}{Proposition}
\newtheorem*{lemma*}{Lemma}
\newtheorem*{definition*}{Definition}
\newtheorem*{condition*}{Condition}

\newtheorem{theorem}{Theorem}
\newtheorem{proposition}{Proposition}
\newtheorem{lemma}{Lemma}
\newtheorem{definition}{Definition}

\newtheorem{remark}{Remark}
\newtheorem{example}{Example}
\newtheorem{assumption}{Assumption}
\newtheorem{corollary}{Corollary}

\DeclareMathOperator*{\argmin}{arg\,min}

\title{Absence of Closed-Form Descriptions for Gradient Flow in Two-Layer Narrow Networks
} 

\author{
  Yeachan Park \\
  Korea Instutite for Advanced Study \\
  \texttt{ychpark@kias.re.kr} \\
}

\begin{document}
\maketitle

\begin{abstract}
In the field of machine learning, comprehending the intricate training dynamics of neural networks poses a significant challenge. This paper explores the training dynamics of neural networks, particularly whether these dynamics can be expressed in a general closed-form solution. We demonstrate that the dynamics of the gradient flow in two-layer narrow networks is not an integrable system. Integrable systems are characterized by trajectories confined to submanifolds defined by level sets of first integrals (invariants), facilitating predictable and reducible dynamics. In contrast, non-integrable systems exhibit complex behaviors that are difficult to predict. To establish the non-integrability, we employ differential Galois theory, which focuses on the solvability of linear differential equations. We demonstrate that under mild conditions, the identity component of the differential Galois group of the variational equations of the gradient flow is non-solvable. This result confirms the system’s non-integrability and implies that the training dynamics cannot be represented by Liouvillian functions, precluding a closed-form solution for describing these dynamics. Our findings highlight the necessity of employing numerical methods to tackle optimization problems within neural networks. The results contribute to a deeper understanding of neural network training dynamics and their implications for machine learning optimization strategies.
\end{abstract}



\section{Introduction}

Gradient-based optimization algorithms have demonstrated remarkable success in addressing non-convex optimization problems within neural networks. However, understanding the training dynamics of these networks remains challenging. How well do we grasp the behavior of parameters during training, and is it feasible to fully comprehend the entire training process? These questions highlight the complexity of neural network optimization and prompt further exploration into its underlying mechanisms.

\begin{figure*}[h]
\begin{center}
   \includegraphics[width=0.9\textwidth]{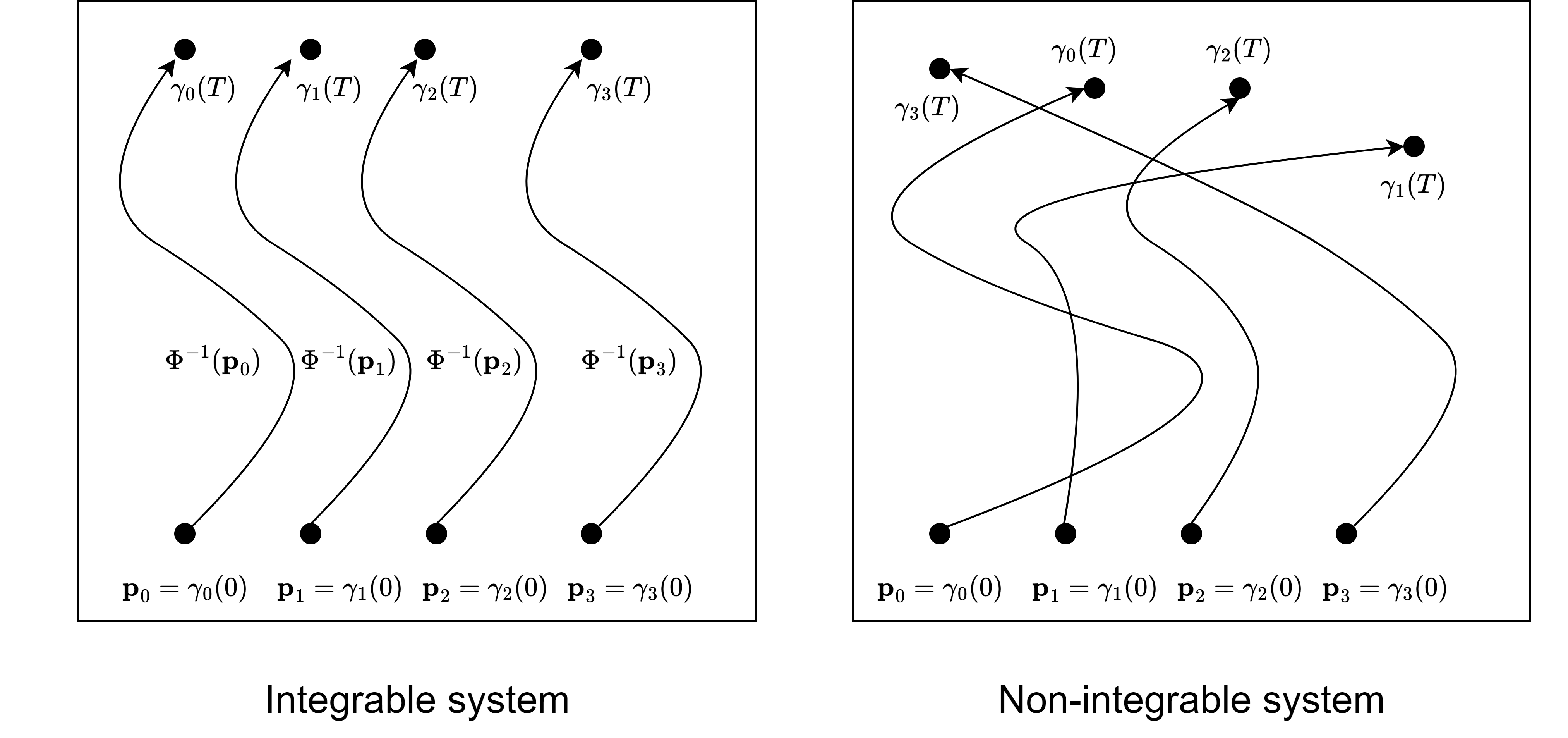}
\end{center}
   \caption{Examples of $2$-dimensional integrable and non-integrable dynamical systems are presented. In the case of the integrable system with the first integral $\Phi$, trajectories are predictable and constrained within the level set of $\Phi$ (left). In contrast, for non-integrable systems, trajectories lack well-defined orderliness (right).  }
\label{fig:dsystem}
\end{figure*}

One approach to assessing the degree of orderliness in a dynamical system is by quantifying the number of conserved quantities within the system. In the context of dynamical system literature, these conserved quantities are often referred to as \emph{first integrals}, which maintain constant over time regardless of the system's evolution. If $n$-dimensional dynamical system has $(n-1)$ first integrals, we say the system is \emph{completely integrable} (Definition \ref{def:cintegralbe}). 
In such cases, the system's trajectory is determined by first integrals and confined within a one-dimensional manifold, which is the intersection of level sets of $(n-1)$ first integrals, thereby enabling predictions of its destination. In a weaker sense, if the dynamical system has $(n-k)$ first integrals and linearly independent $k$ vector fields, it is termed \emph{B-integrable} (Definition \ref{def:bintegrable}). B-integrable system can be viewed as locally completely integrable, suggesting that, they exhibit local predictability due to their locally complete integrability. If the system is integrable with meromorphic first integrals and vector fields, we say the system is integrable \emph{in the meromorphic category} (Definition \ref{def:inte_mero_cat}).
Figure \ref{fig:dsystem} illustrates an example of an integrable system.

So, how can we analyze the complexity of the given dynamical system? To address this question, we introduce the differential Galois theory, a mathematical framework for assessing the solvability of linear differential equations. We say a trajectory with the initial point as the \emph{integral curve} (Definition \ref{def:integralcurve}) of the system. \emph{Variational equations} (Definition \ref{def:vari_eq}) along the integral curve represent linear differential equations describing how perturbations evolve along the trajectory. The \emph{differential Galois group} of the variational equation becomes the key to explore the integrability of the system. 
Let $G^0$ denote the identity component\footnote{ The identity component of a topological group is the connected component that contains the identity element of the group.} of the differential Galois group of the given variational equation.  If $G^0$ is a non-abelian group, the system is not B-integrable in the meromorphic category. Furthermore, if $G^0$ is a non-solvable group, we can prove that no closed-form formula exists to describe the system's complete dynamics (Theorem \ref{theorem:closedform}).

Here, we specify the notion of a closed-form formula throughout the paper.
If a function can be expressed using Liouvillian functions, it can be described in closed-form. Liouvillian functions encompass the majority of elementary functions encountered in calculus.\footnote{For instance, functions such as $x^2,\sqrt{x},e^x, \log(x),\sin(x),\text{erf}(x):=\frac{2}{\sqrt{\pi}}\int_{0}^x e^{-t^2} dt,$ $\text{Ei}(x):=\int_{-\infty}^x \frac{e^t}{t} dt$  are in the category of Liouvillian functions.}  We present an informal definition of this concept. 
\begin{definition*}[informal, Liouvillian function]
   A function $f(x)$ is called Liouvillian if $f(x)$ is representable by a finite numbers of additions, multiplications, $n$-th roots, exponentials, and anti-derivatives.
\end{definition*}
The full definition of Liouvillian  function is presented in \cref{def:liouvillian}. In this paper, we define "closed-form" as being representable by Liouvillian functions.

Then, considering the gradient flow in neural networks, one might ask about complexity of the training dynamics. In this paper, we demonstrate that the gradient flow of neural networks is sufficiently complex (not B-integrable in the meromorphic category) and cannot be expressed in closed-form. To see this, we examine the following simplest two-layer narrow network with ReLU-like smooth activation $\sigma(x)$ (Assumption \ref{assum:1}), $\ell^2$ loss and only four parameters 
$w_1,b_1,w_2,b_2 \in \mathbb{R}$ with the dataset $x_1, ... , x_N$, $y_1, ... , y_N \in \mathbb{R}$. 
\begin{align}
     F(x) &= w_2\sigma(w_1 x+ b_1) + b_2, \label{eq:network} \\ 
     \mathcal{R}  &= \frac{1}{2} \sum_{i=1}^N (F(x_i) -y_i)^2 = \frac{1}{2} \sum_{i=1}^N (w_2\sigma(w_1 x_i+ b_1) + b_2 -y_i)^2,
\end{align}
where $N$ denotes the number of samples.

In this scenario, we demonstrate that the gradient flow of even such a simple network \cref{eq:network} is complex enough (not B-integrable in the meromorphic category). This is established through the application of the differential Galois theory. The proof proceeds as follows: initially, we find the integral curve of the gradient flow. Next, we derive the variational equation associated with this curve. Finally, by showing that the differential Galois group of the variational equation is non-abelian, we establish the non-integrability of the gradient flow.

Furthermore, we establish that there exists no closed-form expression (Liouvillian expression) to describe the complete dynamics of the gradient flow of network \cref{eq:network}. This is demonstrated by showing that the differential Galois group is non-solvable.  While there have been works to exactly solve the gradient flow of deep linear networks \cite{saxe2014exact,saxe2019mathematical}  or matrix factorization problems \cite{tarmoun2021understanding}, our findings suggest the impossibility of such exact closed-form solutions for nonlinear neural networks. 
 This result suggests that solving the gradient flow in an explicit form, as demonstrated in simple linear regression problems (Equation \eqref{eq:linearmodel}), is not possible.  Instead, numerical iterative methods become indispensable for solving the gradient flow.
\begin{align}
    &L(W) = \frac{1}{2}\| X^T W - y \|^2, \; \hat{W} = \argmin_W L(W) = (XX^T)^{-1}Xy \label{eq:linearmodel},  \nonumber \\
    &W(t) = W(0)- (XX^T)^{-1} X (I-\exp(-X^T X t))(X^T W(0) - y ). 
\end{align}
\begin{figure*}[h]
\begin{center}
   \includegraphics[width=0.9\textwidth]{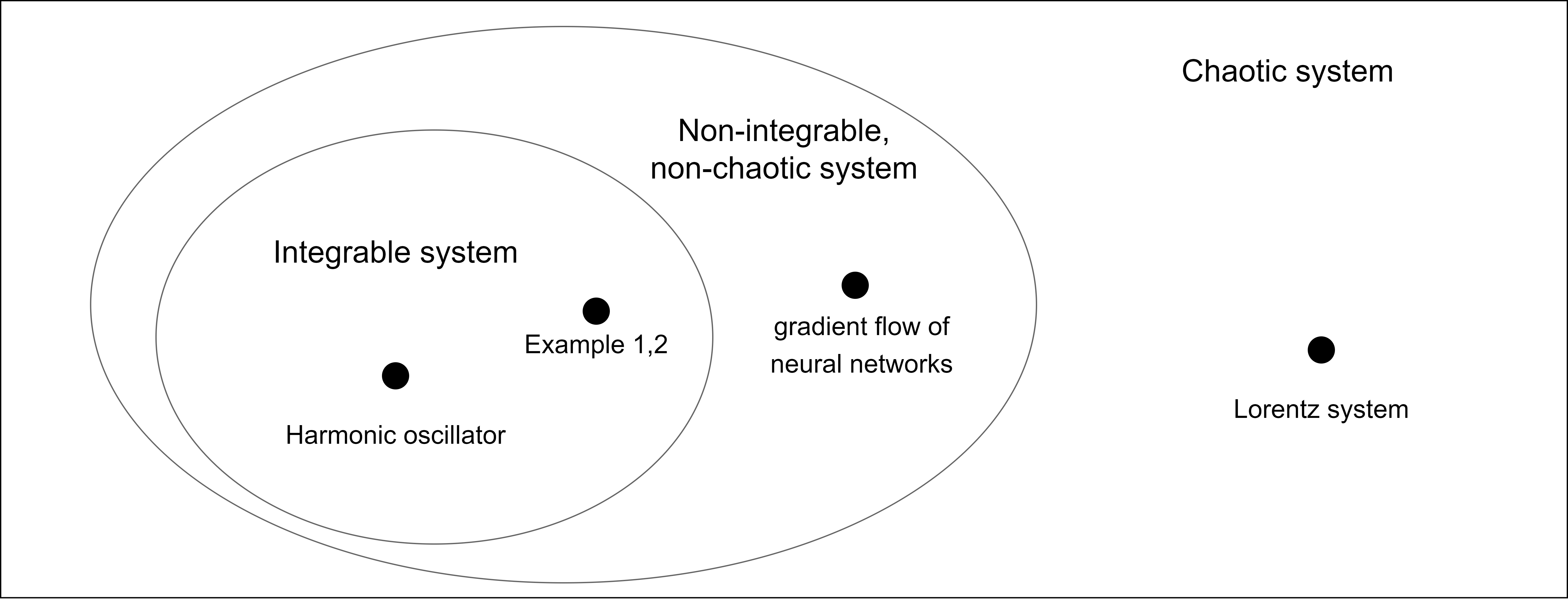}
\end{center}
   \caption{A diagram of various dynamical systems. In this diagram, we denote an integrable system as an integrable system in the meromorphic category. }
\label{fig:diagram_system}
\end{figure*}

Nevertheless, it is important to note that non-integrability does not necessarily imply chaos in training neural networks.  When the activation function is analytic and the gradient flow possesses a limit point, convergence is guaranteed by Łojasiewicz inequality \cite{colding2014lojasiewicz}. Additionally, a wealth of research also supports the convergence of both gradient flow and gradient descent in sufficiently wide neural networks \cite{chizat2018global,du2019gradient,du2018gradient,karimi2016linear}. Thus, we assert that the gradient flow of a neural network is non-integrable and non-chaotic, positioning it between integrable and chaotic systems (Figure \ref{fig:diagram_system}). 
Now, we present our contributions as follows.
\begin{itemize}
    \item We show that under a mind condition (Assumption \ref{assum:1}, $\sum_i x_i \ne 0$), the gradient flow of even a two-layer narrow network is not integrable 
 (Theorem \ref{theorem:main}). 
 \item In addition, we show that full dynamics of gradient flow cannot be expressed by closed-form (Theorem \ref{theorem:closedform}, Corollary \ref{cor:main}).
\end{itemize}

\section{Related Work}
\subsection{Integrability of dynamical systems}
The classical three-body problem has played a significant role in the development of integral systems theory.  
Morales-Ruiz, Ramis, and Simo formulate a theory that describes the meromorphic integrability of Hamiltonian systems using differential Galois theory \cite{morales2001galoisian,morales2007integrability}. Subsequently, \cite{tsygvintsev2001meromorphic} and \cite{boucher2003application} prove the non-integrability of the planar three-body problem.

\cite{ayoul2010galoisian} extend the Morales-Ramis theory to encompass non-Hamiltonian dynamical systems. 
\cite{acosta2018differential} introduce a novel technique for assessing the integrability of analytic planar vector fields. Leveraging this approach, \cite{llibre2005formal}, \cite{huang2018meromorphic}, and \cite{yang2021first} examine  the non-integrability of Lorentz system.   \cite{yagasaki2023nonintegrability} prove the non-integrability of epidemic SEIR models.

\subsection{Difficulties in training dynamics of neural networks}
There are several studies on the difficulties of training neural networks. In terms of computational complexity, \cite{blum1988training} demonstrate that determining the weights of 3-node networks with threshold activations is NP-hard.  \cite{jones1997computational} establishes the NP-hardness of training 3-node networks with sigmoid activations.  \cite{goel2021tight} and \cite{dey2020approximation} prove that training 2-layer ReLU networks is also NP-hard.

In the context of chaotic systems, \cite{sasdelli2021chaos} empirically demonstrate that continuous training dynamics using SGD exhibit locally chaotic behavior. Similarly, \cite{herrmann2022chaotic} empirically show that time-discrete dynamical systems trained by SGD also entail locally chaotic behavior and propose a non-chaotic modified SGD. It is noteworthy that  \cite{sasdelli2021chaos} and \cite{herrmann2022chaotic} focus on \textit{local} chaos as evidenced by the presence of positive local Lyapunov exponents. Specifically, \cite{herrmann2022chaotic} demonstrate that \textit{global} chaos emerges at the beginning of the training, and diminishes towards the end of training.

\section{Preliminaries}
\label{section:prelim}
In this paper, we assume that the dynamical systems are in $\mathbb{C}$ (the set of complex numbers) for generality. We observe that the actual dynamical systems are projections onto $\mathbb{R}$ (real numbers) of the systems defined in $\mathbb{C}$.

\subsection{Integrable dynamical system}
Consider the following $n$-dimensional dynamical system with vector field $F(\bold{x})$
\begin{align}
    \frac{d \bold{x}(t)}{dt} = F( \bold{x} ), \; \bold{x}(t) \in \mathcal{M}, \; t \in \mathbb{C}, \label{eq:ds}
\end{align}
where $F(\cdot) : \mathbb{C}^n \to \mathbb{C}^n$ is an $n$-dimensional analytic vector-valued function and $\mathcal{M}$ is an $n$-dimensional complex manifold.

An integral curve of the dynamical system represents the trajectory of the system's state over time, given its initial point. 
\begin{definition}[integral curve]
\label{def:integralcurve}
     We say $\gamma: \mathbb{C} \to \mathcal{M}$ is an integral curve of dynamical system \eqref{eq:ds} if it follows dynamical system \eqref{eq:ds} with the initial point $\bold{p} \in \mathcal{M}$:
    \begin{align}
        \frac{d \gamma(t) }{dt} = F(\gamma(t)), \; \gamma(0) = \bold{p},
    \end{align}
 for $t \in \mathbb{C}$.
\end{definition}

A first integral of the dynamical system is a function that remains constant along the trajectories of the system. First integrals are also refereed as conserved quantities or invariants of the system.
\begin{definition}[first integral]
\label{def:firstintegral}
     We say $\Phi : \mathbb{C}^n \to \mathcal{M}$ is a first integral of dynamical system \eqref{eq:ds} if 
    \begin{align}
        \frac{d \Phi(\gamma(t)) }{dt} = 0, \; \forall t \in \mathbb{C},
    \end{align}
for any integral curve $\gamma(t)$ of dynamical system \eqref{eq:ds}. 
\end{definition}

\begin{definition}[functionally independent]
    Let $f_1, ..., f_k : \mathcal{M} \to \mathbb{C}$ be $n$ smooth functions where $\mathcal{M}$ is $n$-dimensional complex manifold. We say $f_1,..., f_k$ are functionally independent if $\nabla f_1 , ... , \nabla f_k$ are linearly independent, \textit{i.e.}, if the Jacobian has full rank. 
\end{definition}
If $f_1, ... , f_k$ are functionally independent, then their level sets intersect transversely. 

\begin{definition}[completely integrable]
\label{def:cintegralbe}
    We say system \eqref{eq:ds} is completely integrable if it admits $(n-1)$ functionally independent first integrals $\Phi_1, ... , \Phi_{n-1}$. 
\end{definition}

If system \eqref{eq:ds} is completely integrable, then its trajectory $\gamma(t)$ is analytically obtained in an explicit way since $\gamma(t)$ is contained in specific one-dimensional manifold $\mathcal{N}$,
\begin{align}
    \gamma(t) \subset \mathcal{N}, \; \mathcal{N}=\bigcap_{i=1}^{n-1} \{ p \in \mathcal{M} : \Phi_i(p) = c_i, \; c_i \in \mathbb{C} \},
\end{align}
and $c_i$'s are specified by an initial point $\gamma(0)$.

\begin{definition}[B-integrable]
\label{def:bintegrable}
    We say system \eqref{eq:ds} is B-integrable if it admits $(n-k)$ functionally independent first integrals $\Phi_1, ... , \Phi_{n-k}$ and linearly independent $k$ vector fields $X_1=F, X_2, ... , X_{k}$ such that 
    \begin{align}
        [X_i, X_j] = 0, \; 1 \le i \le j \le k \; \text{and} \; X_i(\Phi_j) = 0, \; 1 \le i \le k, \; 1 \le j \le n-k, 
    \end{align}
where $[X_i, X_j](\cdot) = X_i(X_j(\cdot))-X_j(X_i(\cdot))$ is the Lie bracket. 
\end{definition}
The notion of B-integrability was introduced by \cite{bogoyavlenskij1998extended}. 
If system \eqref{eq:ds} is B-integrable, $n-k$ functionally independent first integrals $\Phi_1, ..., \Phi_{n-k}$ form a $k$-dimensional level set $\mathcal{N}$ which is $k$-dimensional submanifold $\mathcal{N} = \bigcap_{i=1}^{k} \{ p \in \mathcal{M} : \Phi_i(p) = c_i, \; c_i \in \mathbb{C} \} \subset \mathcal{M}$. The trajectory $\gamma(t)$ lies on $\mathcal{N}$.    Additionally, since $k$ linearly independent and commuting vector fields $X_1=F, ..., X_{k}$ complete  $\mathcal{N}$, for each $p \in \mathcal{N}$ there exists a local coordinate chart $(U,\phi)$ centered at $p$ with $\phi:(x_1,...,x_n) \mapsto (s_1, ... , s_k)$  such that $X_i = \frac{\partial}{\partial s_i}$ for $i=1,...,k$.\footnote{Theorem 9.46 in \cite{lee2003introduction}.} Therefore, system \eqref{eq:ds} is completely integrable in $U \subset \mathcal{N}$ with $(k-1)$ first integrals $\Psi_2:=\phi^{-1}(s_2) , ... , \Psi_k := \phi^{-1}(s_k)$. Hence, if the system is B-integrable, then we can conclude that the system is locally completely integrable. 

\begin{definition}[meromorphic function]
A function $f:\mathcal{M} \subset \mathbb{C}^n \to \mathbb{C}$ is called meromorphic on $\mathcal{M}$ if for every $a \in \mathcal{M}$, there is a neighborhood $U \ni a$ and holomorphic functions $p(z),q(z)$ where $q(z)$ is not identically $0$ on $U$ such that 
\begin{align}
    f(z)  = \frac{p(z)}{q(z)}, \; z \in U \setminus q^{-1}(0).
\end{align}
\end{definition}
In other words, a meromorphic function is a function that can be expressed as the quotient of holomorphic functions.
\begin{definition}[integrable in the meromorphic category]
\label{def:inte_mero_cat}
    If system \eqref{eq:ds} is completely integrable and the vector field $X(t)$ and $(n-1)$ functionally independent first integrals are meromorphic, then we say the system is completely integrable in the meromorphic category. \\
    If system \eqref{eq:ds} is B-integrable and $(n-k)$ functionally independent first integrals and $k$ commuting vector fields are meromorphic, we say the system is B-integrable in the meromorphic category.
\end{definition}

\subsubsection{Examples of dynamical systems}

We present some examples of integrable dynamical systems.
\begin{example}
\label{ex:1}
    Let $\mathcal{M} = \mathbb{C}^2$. Consider the $2$-dimensional dynamical system with the vector field $F(\bold{x}) = (-y,x), \; \bold{x}=(x,y) \in \mathbb{C}^2:$
\begin{align}
    \frac{d \bold{x}(t)}{dt} = \begin{bmatrix}
        x' \\
        y'
    \end{bmatrix}= \begin{bmatrix}
        -y \\
        x
    \end{bmatrix}.
\end{align}
Then the system is completely integrable in the meromorphic category since  it has a meromorphic first integral $\Phi_1(x,y) = x^2+y^2$. Hence, the trajectory of the system has an explicit form of $\gamma(t)= (x_0\cos(t)-y_0 \sin(t),x_0\sin(t)+y_0\cos(t))$ with the initial point $\gamma(0)=(x_0,y_0) \in \mathbb{C}^2$.   
\end{example}

\begin{example}
\label{ex:1_1}
    Let $\mathcal{M} = \mathbb{C}^2$. Consider the $2$-dimensional dynamical system with the vector field $F(\bold{x}) = (x,y), \; \bold{x}=(x,y) \in \mathbb{C}^2:$
\begin{align}
    \frac{d \bold{x}(t)}{dt} = \begin{bmatrix}
        x' \\
        y'
    \end{bmatrix}= \begin{bmatrix}
        x \\
        y
    \end{bmatrix}.
\end{align}
Then the system is completely integrable in the meromorphic category since it has a meromorphic first integral $\Phi_1(x,y) =\frac{y}{x}.$ 
Hence, the trajectory of the system has an explicit form of $\gamma(t)= (x_0 e^t , y_0 e^t)$ with the initial point $\gamma(0)=(x_0,y_0) \in \mathbb{C}^2$.   
\end{example}

\begin{example}[simple harmonic oscillator]
\label{ex:harmonicO}
We consider a motion of a simple harmonic oscillator. 
\begin{align}
    \frac{d^2 x(t)}{dt^2} + \omega_0^2 x(t)=0, \label{eq:harmonicO}
\end{align}
where $x(t)$ is the displacement of the oscillator and $ \omega_0 \ne 0$ is the angular frequency of the oscillator. We can restate equation \eqref{eq:harmonicO} as a dynamical system.
\begin{align}
    \frac{d \bold{x}(t)}{dt} = \begin{bmatrix}
        x' \\
        y'
    \end{bmatrix}= \begin{bmatrix}
        y \\
        -\omega_0^2 x
    \end{bmatrix}, \label{eq:ds_hos}
\end{align}
where $\bold{x}(t)=(x(t),y(t))$, $y(t):=x'(t)$ is the velocity of the oscillator. The trajectory of the system has an explicit form of $\gamma(t) = (\frac{v_0}{\omega_0}\sin(\omega_0 t)+ x_0 \cos(\omega_0 t), v_0\cos(\omega_0 t)- \omega_0 x_0 \sin(\omega_0 t) )$ with the initial condition $x(0)=x_0,x'(0)=v_0$. Moreover, the system is completely integrable in the meromorphic category since it has a meromorphic first integral $\Phi(x,y)=\omega_0^2x^2+y^2$.
\end{example}

Next, we present the famous Lorentz system \cite{lorenz1963deterministic}.
\begin{example}[Lorentz system]
    Let $\mathcal{M}= \mathbb{C}^3$. Consider the Lorentz system 
    \begin{align}
        x'(t) &= - \sigma x + \sigma y \\
        y'(t) &= r x -y - xz \\
        z'(t) &= xy - \beta z,
    \end{align}
where $\sigma,r,\beta \in \mathbb{C}$. \\
If $\sigma=0$, \cite{llibre2005formal} show that the Lorentz system is completely integrable with two first integrals.
\cite{huang2018meromorphic} show that if $\sigma\beta \ne 0$, the Lorentz system is not completely integrable in the meromorphic category. Moreover if $\frac{2\sqrt{(\sigma+1)^2+4\sigma(r-1)}}{|\beta|}$ is not odd integer, the Lorentz system is not B-integrable in the meromorphic category. Moreover, the Lorentz system is chaotic for specific configurations of $(\sigma,r,\beta)$ \cite{mischaikow1995chaos,mischaikow1998chaos,mischaikow2001chaos}.
\end{example}

\subsection{Differential Galois theory}
Differential Galois theory plays an important role in demonstrating the impossibility of representing solutions of linear ODEs in closed-form. In this section, we present the formal definition of Liouvillian functions. We defer a brief introduction to differential Galois theory to \cref{section:differentialGalois}.

\begin{definition}[Liouvillian extension, Liouvillian function]
\label{def:liouvillian}
    Let $K \subset L$ be a differential field extension. We say $l \in L$ is Liouvillian over $K$ if $l$ is either algebraic, primitive, or exponential over $K$. Similarly, a differential field extension $K \subset L$ is Liouvillian if there exists a finite sequence of
intermediate differential field extensions 
\begin{equation}
    K = K_0 \subset K_1 \subset ... \subset K_n = L,
\end{equation}
such that $K_{i+1} = K_i(l_i)$ and $l_i$ is Liouvillian over $K_i \; $ for $1 \le i \le n$. 
$l$ is called Liouvillian over $K$ if $K \subset K(l)$ is a Liouvillian extension.  \\
If $l$ is Liouvillian over $\mathbb{C}(x)$, we simply call $l$ be a Liouvillian function. 
\end{definition}
Thus, if $l$ is Liouvillian, $l$ is representable by a finite number of additions, multiplications, $n$-th roots, exponentials, and anti-derivatives of algebraic functions. 

Next, we present the useful lemma to determine the solvability of specific form of the second order ODE.
\begin{lemma}
\label{lemma:2ndode}
    Consider the following second order ODE
\begin{align*}
    L(y) &= y(t)'' - r y(t), \\
    r &= a_2 t^2 + a_1 t + a_0 + \frac{a_{-1}}{t+d}+ \frac{b}{(t+d)^2},
\end{align*}
where $a_2,a_1,a_0,a_{-1},b,d \in \mathbb{C}$. If $a_2, b\ne 0$ and  $-2+\sqrt{1+4b}$ is not a nonnegative integer,  then $L(y)$ has no Liouvillian solution and its differential Galois group is $Gal(L(y)) = SL_2(\mathbb{C})$. 
\end{lemma}
\begin{proof}
     The proof is presented in \cref{proof:2ndode}.
\end{proof}

\subsection{Morales-Ramis theory on integrability of dynamical systems}
Now we present the Morales-Ramis theory, an application of differential Galois theory to dynamical systems, which describes the integrability of these systems.

\begin{figure*}[h]
\begin{center}
   \includegraphics[width=0.7\textwidth]{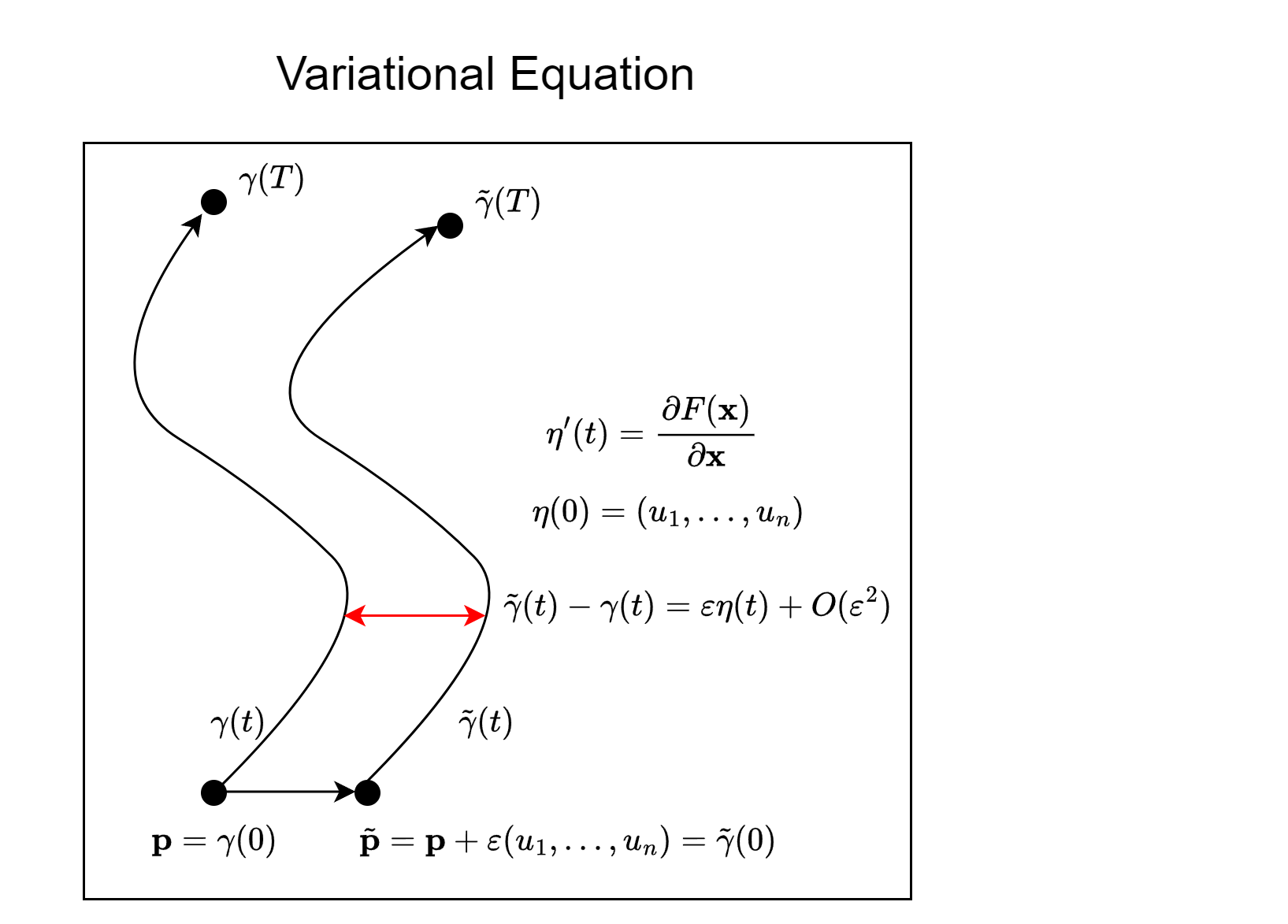}
\end{center}
   \caption{An illustration of variational equations of dynamical system $ \frac{d \bold{x}(t)}{dt} = F( \bold{x} )$. For the integral curve $\gamma(t)$ with the initial point $\bold{p}$, $\tilde{\gamma}(t)$ denotes the perturbed curve with the perturbed initial point $\tilde{\bold{p}}$. Let $\eta(t)$ follow variational equation \eqref{eq:ves} with the initial condition $\eta(0)=(u_1,...,u_n)$. Then $\varepsilon \eta(t)$ is the first-order approximation (in terms of $\varepsilon$) of the perturbations ($\tilde{\gamma}(t)-\gamma(t)$) of the integral curve $\gamma(t)$. }
\label{fig:ve}
\end{figure*}

\begin{definition}[variational equation]
\label{def:vari_eq}
    Consider dynamical system \eqref{eq:ds}. Let $\gamma(t)$ be an integral curve of the system. 
Then we consider the following linear differential equation for $\eta(t) : \mathbb{C} \to \mathbb{C}^n$:
    \begin{equation}
        \frac{d \eta(t) } {dt } = A(t) \eta(t), \; A(t) = \frac{\partial F(\bold{x}(t))}{\partial \bold{x}} \in \mathbb{C}^{n \times n}, \; \eta(t) \in T \mathcal{M}, \label{eq:ves}
    \end{equation}
    where $T \mathcal{M}$ is the tangent bundle. Equation \eqref{eq:ves} is called the variational equation ($VE_{\gamma(t)}$) along $\gamma(t)$. 
\end{definition}
The variational equation describes how perturbation evolves along the trajectory $\gamma(t)$. Please refer to Figure \ref{fig:ve} for an illustration of variational equations.
\begin{remark}
\label{remark:ve}
We can understand variational equations as follows. Let $\gamma(t)$ be an integral curve of the dynamical system \eqref{eq:ds} with the initial point $\mathbf{p}$. Let $\tilde{\mathbf{p}}=\mathbf{p}+\varepsilon( u_1, ... , u_n)$ be a perturbed point of $\mathbf{p}$ with a small perturbation $\varepsilon$, and let $\tilde{\gamma}$ be the integral curve with initial point $\tilde{\mathbf{p}}$. Then the difference $\tilde{\gamma}(t) - \gamma(t)$ follows the equation
\begin{align}
    \tilde{\gamma}(t) - \gamma(t) = \varepsilon \eta(t) + O(\varepsilon^2), \; \text{as} \; \varepsilon \to 0, \label{eq:ved}
\end{align}
where $\eta(t)$ satisfies 
\begin{align*}
    \eta'(t) &= \lim_{\varepsilon \to 0 } \frac{F(\tilde{\gamma}(t))-F(\gamma(t)) }{\varepsilon} \\
    &=  \lim_{\varepsilon \to 0 } \frac{F( \gamma(t)+ \varepsilon \eta(t)+ O(\varepsilon^2) )-F(\gamma(t)) } {\varepsilon} = \frac{\partial F(\bold{x})}{\partial \bold{x}} \eta(t), 
\end{align*}
with the initial condition $\eta(0) = (u_1, ... , u_n)$. Therefore $\eta(t)$ is the solution of the variational equation. Hence, we can conclude that the variational equation describes the first-order approximation (in terms of $\varepsilon$) of the perturbation of the integral curve $\gamma(t)$. 
\end{remark}
Now, we present an important statement to demonstrate the non-integrability of the dynamical systems.
\begin{lemma}[\cite{ayoul2010galoisian}]
\label{lemma:ayoul}
    Suppose a dynamical system is B-integrable in the meromorphic category in a neighborhood of its integral curve $\gamma(t)$. Then the identity component $Gal^0(VE_{\gamma(t)})$ of the differential Galois group $Gal(VE_\gamma)$ of the variational equations $VE_{\gamma(t)}$ is abelian. 
\end{lemma}
For a given linear differential equation $L(y)$, the differential Galois group $Gal(L(y))$ is the group of automorphisms that map the solutions of $L(y)$ to themselves. We can demonstrate the non-integrability of the system by showing that $Gal^0(VE_{\gamma(t)})$ is not abelian. The integrabililty of the dynamical system is closely related to the differential Galois group. \cref{lemma:ayoul} is useful for determining the integrability of the dynamical system. We will show the non-integrability of the gradient flow of neural networks using \cref{lemma:ayoul}.

We provide a brief history of Morales-Ramis theory. Morales, Ramis and Simo demonstrate the obstruction to meromorphic integrability of Hamiltonian systems in terms of differential Galois theory \cite{morales2001galoisian,morales2007integrability}. \cite{ayoul2010galoisian} further extend this result to the general non-Hamiltonian dynamical system.

\section{Non-integrability of gradient flow}
In this section, we present our main statement that the gradient flow of a two-layer narrow network is not B-integrable in the meromorphic category.
First, we assume that the activation function $\sigma(x)$ satisfies the following assumption:
\begin{assumption}
\label{assum:1}
    Suppose that the activation function $\sigma(x)$ is analytic and $\sigma'(\hat{b}) = 0$ at some $\hat{b} \in \mathbb{R}$ with $\sigma(\hat{b}), \sigma''(\hat{b}) \ne 0$. 
\end{assumption}
Roughly speaking, $\sigma(x)$ locally exhibits a non-zero U-like shape around around $x=\hat{b}$. 
Modern smooth activation functions feature a smoother ReLU-like profile ($\sigma(x) \to 0$ as $x \to -\infty$ and $\sigma(x) \to x$ as $x \to \infty$). Consequently, most smooth activation functions satisfy Assumption \ref{assum:1}. Notably, a variety of activation functions such as \texttt{SiLU} \cite{elfwing2018sigmoid}, \texttt{SoftPlus}, \texttt{GELU} \cite{hendrycks2016gaussian}, \texttt{Swish} \cite{ramachandran2017searching}, and \texttt{Mish} \cite{misra2019Mish} satisfy Assumption \ref{assum:1}. 

Now, we restate the following two-layer narrow network with $\ell^2$ loss and four parameters 
$W=(w_1,b_1,w_2,b_2) \in \mathbb{R}^4$ with the dataset $x_1, ... , x_N, y_1, ... , y_N \in \mathbb{R}$. 
\begin{align*}
     F(x,W) &= w_2\sigma(w_1 x+ b_1) + b_2,  \\ 
     \mathcal{R}(W)  &= \frac{1}{2} \sum_{i=1}^N (F(x_i,W) -y_i)^2 = \frac{1}{2} \sum_{i=1}^N (w_2\sigma(w_1 x_i+ b_1) + b_2 -y_i)^2,
\end{align*}
where $N$ denotes the number of samples. $F(x,W)$ denote the output of the network with input $x$ and the set of parameters $W$, and $\mathcal{R}(W)$ denote the $\ell^2$ loss. 
Then, we have the following dynamical system of the gradient flow:
\begin{align*}
    W'(t) = - \frac{\partial \mathcal{R}}{\partial W}.
\end{align*}
The component-wise expressions of the gradient flow are as follows:
\begin{align}
    w_1'(t) &= -\frac{\partial \mathcal{R}}{\partial w_1} = -\sum_{i=1}^N (w_2\sigma(w_1 x_i+ b_1) + b_2 -y_i) (w_2 \sigma'(w_1 x_i + b_1) ) x_i, \label{eq:gf1}\\
    b_1'(t) &= -\frac{\partial \mathcal{R}}{\partial b_1} = -\sum_{i=1}^N (w_2\sigma(w_1 x_i+ b_1) + b_2 -y_i) (w_2 \sigma'(w_1 x_i + b_1) ), \label{eq:gf2}\\
    w_2'(t) &= -\frac{\partial \mathcal{R}}{\partial w_2} = -\sum_{i=1}^N (w_2\sigma(w_1 x_i+ b_1) + b_2 -y_i) ( \sigma(w_1 x_i + b_1) ),\label{eq:gf3}\\
    b_2'(t) &= -\frac{\partial \mathcal{R}}{\partial b_2} = -\sum_{i=1}^N (w_2\sigma(w_1 x_i+ b_1) + b_2 -y_i). \label{eq:gf4}
\end{align}
Now we formally present our main theorem. 
\begin{theorem}
\label{theorem:main}
    Suppose the activation function $\sigma$ satisfies Assumption \ref{assum:1} and 
 \\ $ \sum_{i=1}^N x_i \ne 0$. 
    Then the dynamical system of the gradient flow \eqref{eq:gf1}-\eqref{eq:gf4} is not B-integrable in the meromorphic category. 
\end{theorem}
\begin{remark}
    We clarify that throughout the proof, we consider all numbers as belonging to $\mathbb{C}$ although they are actually real numbers as mentioned in Section \ref{section:prelim}. 
\end{remark}

\begin{proof}
Before beginning the proof, we will briefly outline our proof strategy.
First, we find the integral curve $\gamma(t)$ of the gradient flow (\cref{lemma:icurve}). Next, we obtain the variational equation $\eta(t)$ along the integral curve $\gamma(t)$ (\cref{lem:vari})). Then we reduce the variational equation $\eta(t)$ into a second-order ODE (\cref{lemma:epw2,lemma:28,lemma:tau}). Using \cref{lemma:2ndode}, we verify that the variation equation has no Liouvillian solution and that its differential Galois group is not abelian. Finally using \cref{lemma:ayoul}, we conclude that gradient flow is not B-integrable in the meromorphic category. The proofs of \cref{lemma:icurve,lem:vari,lemma:epw2,lemma:28,lemma:tau} are presented in \cref{sec:proof_lemmas}.

Now, we begin the proof by defining the following five quantities.
\begin{align*}
    \overbar{x} &= \frac{1}{N} \sum_{i=1}^N x_i , \; \overbar{y} = \frac{1}{N} \sum_{i=1}^N y_i, \; \overbar{x^2} = \frac{1}{N}\sum_{i=1}^N x_i^2 ,   \\
    \overbar{xy} &= \frac{1}{N}\sum_{i=1}^N x_iy_i, \; \overbar{x^2y } = \frac{1}{N}\sum_{i=1}^N x_i^2y_i.
\end{align*}
Next, we find the following integral curve $\gamma(t)$ starting from an initial point $\gamma(0)=(0,\hat{b},a \sigma(\hat{b}),\hat{b}_2)$: 
\begin{align}
   \gamma(t) = (w_1(t), b_1(t), w_2(t), b_2(t)) = (0,\hat{b},a \sigma(\hat{b}) e^{-\omega t}, a e^{-\omega t }+\overbar{y}  ), \label{eq:ic}
\end{align}
where $\omega = N(\sigma(\hat{b})^2+1),  a = \hat{b}_2 - \overbar{y} \ne 0$, and $\hat{b}_2 \in \mathbb{R}$ can be arbitrary and determined later. 
\begin{lemma}
\label{lemma:icurve}
    $\gamma(t)$ is an integral curve of the gradient flow \eqref{eq:gf1}-\eqref{eq:gf4}. 
\end{lemma}
To exploit Lemma \ref{lemma:ayoul}, we attain the variational equation along the integral curve $\gamma(t)$.  
We have the following variational equation $\eta(t)$ along $\gamma(t)$:
\begin{align}
    &\frac{d \eta(t)}{dt} = \frac{\partial}{\partial W} (-\frac{\partial \mathcal{R}}{\partial W} ) \eta(t) = \frac{\partial}{\partial W} \left( \begin{bmatrix}  {w'_1}(t) \\ {b'_1}(t) \\ {w'_2}(t)  \\ {b'_2}(t)  \end{bmatrix}  \right) \begin{bmatrix}  \epsilon_{w_1}(t) \\ \epsilon_{b_1}(t) \\ \epsilon_{w_2}(t)  \\ \epsilon_{b_2}(t)    \end{bmatrix}, \nonumber \\
    & \eta'(t) = \begin{bmatrix}   \epsilon_{w_1}'(t) \\ \epsilon_{b_1}'(t) \\ \epsilon_{w_2}'(t)  \\ \epsilon_{b_2}'(t)    \end{bmatrix}  = \begin{bmatrix}  \frac{\partial w_1'(t) } {\partial w_1} & \frac{\partial w_1'(t) } {\partial b_1} & \frac{\partial w_1'(t) } {\partial w_2}  & \frac{\partial w_1'(t) } {\partial b_2} \\
    \frac{\partial b_1'(t) } {\partial w_1} & \frac{\partial b_1'(t) } {\partial b_1} & \frac{\partial b_1'(t) } {\partial w_2}  & \frac{\partial b_1'(t) } {\partial b_2} \\ \frac{\partial w_2'(t) } {\partial w_1} & \frac{\partial w_2'(t) } {\partial b_1} & \frac{\partial w_2'(t) } {\partial w_2}  & \frac{\partial w_2'(t) } {\partial b_2} \\
    \frac{\partial b_2'(t) } {\partial w_1} & \frac{\partial b_2'(t) } {\partial b_1} & \frac{\partial b_2'(t) } {\partial w_2}  & \frac{\partial b_2'(t) } {\partial b_2} \end{bmatrix} \begin{bmatrix}  \epsilon_{w_1}(t) \\ \epsilon_{b_1}(t) \\ \epsilon_{w_2}(t)  \\ \epsilon_{b_2}(t)    \end{bmatrix}. \label{eq:ve_matrix}
\end{align}
for $\eta(t) = (\epsilon_{w_1}(t) , \epsilon_{b_1}(t),  \epsilon_{w_2}(t), \epsilon_{b_2}(t) )^T \in \mathbb{C}^4$.
\begin{lemma}
\label{lem:vari}
The variational equation $\eta(t)$ along the integral curve $\gamma(t)$ has the following forms. 
\begin{align}
     \epsilon_{w_1}'(t) &=  \begin{bmatrix}  -\sum_{i=1}^N   ( a \frac{\omega}{N} e^{-\omega t}  +\overbar{y}- y_i ) (a \sigma(\hat{b}) e^{-\omega t} \sigma''(\hat{b}) x_i^2 )   \\ 
    -\sum_{i=1}^N   ( a \frac{\omega}{N} e^{-\omega t}  +\overbar{y}- y_i ) (a \sigma(\hat{b}) e^{-\omega t} \sigma''(\hat{b}) x_i )   \\ 
0 \\ 
    0    \end{bmatrix}^T 
    \begin{bmatrix}  \epsilon_{w_1}(t) \\ \epsilon_{b_1}(t) \\ \epsilon_{w_2}(t)  \\ \epsilon_{b_2}(t)    \end{bmatrix}, \label{eq:ve1} \\
    \epsilon_{b_1}'(t)  &= \begin{bmatrix} -\sum_{i=1}^N   ( a \frac{\omega}{N} e^{-\omega t}  + \overbar{y}- y_i ) (a \sigma(\hat{b}) e^{-\omega t} \sigma''(\hat{b}) x_i )   \\ 
   -\sum_{i=1}^N   ( a \frac{\omega}{N} e^{-\omega t}  +\overbar{y}- y_i ) (a \sigma(\hat{b}) e^{-\omega t} \sigma''(\hat{b})  )  \\ 
     0  \\ 
    0
    \end{bmatrix}^T  \begin{bmatrix}  \epsilon_{w_1}(t) \\ \epsilon_{b_1}(t) \\ \epsilon_{w_2}(t)  \\ \epsilon_{b_2}(t)    \end{bmatrix}, \label{eq:ve2} \\ 
    \epsilon_{w_2}'(t)  & = \begin{bmatrix}  0    \\ 
   0    \\ 
    - \sum_{i=1}^N  \sigma(\hat{b})^2 \\ 
    - \sum_{i=1}^N  \sigma(\hat{b})    \end{bmatrix}^T \begin{bmatrix}  \epsilon_{w_1}(t) \\ \epsilon_{b_1}(t) \\ \epsilon_{w_2}(t)  \\ \epsilon_{b_2}(t)    \end{bmatrix}, \label{eq:ve3} \\ 
    \epsilon_{b_2}'(t)  &= \begin{bmatrix}  0   \\ 
       0     \\ 
    - \sum_{i=1}^N  \sigma(\hat{b}) \\ 
    - \sum_{i=1}^N 1   \end{bmatrix}^T  \begin{bmatrix}  \epsilon_{w_1}(t) \\ \epsilon_{b_1}(t) \\ \epsilon_{w_2}(t)  \\ \epsilon_{b_2}(t)    \end{bmatrix} . \label{eq:ve4}
\end{align}
\end{lemma}

Therefore, we can separate the case of $\epsilon_{w_1}(t),\epsilon_{b_1}(t)$ and the case of $\epsilon_{w_2}(t),\epsilon_{b_2}(t)$.
\begin{lemma}
\label{lemma:epw2}
    The differential Galois groups for $\epsilon_{w_2}(t)$ and $\epsilon_{b_2}(t)$ are $\mathbb{C}^\ast$, hence they are abelian.
\end{lemma}
Therefore, we only need to care the case of $\epsilon_{w_1}(t),\epsilon_{b_1}(t)$.

\begin{lemma}
\label{lemma:28}
    Let $\tau := e^{- \omega t}$ and $ g_{w_1}(\tau) := \epsilon_{w_1}(t)$. THen we have the following second-order ODE for $g_{w_1}(\tau)$: 
\begin{align}
    g_{w_1}''(\tau) = P_1(\tau) g_{w_1}'(\tau) + P_2(\tau)g_{w_1}(\tau), \label{eq:28}
\end{align}
where
\begin{align*}
    & P_1(\tau) =  A_{1,1}\tau +B_{1,1}+ (A_{1,2}\tau + B_{1,2})^{-1} \big(A_{1,2}+(A_{1,2}\tau +B_{1,2})(A_{2,2}\tau) \big),  \\
    &P_2(\tau) = A_{1,1}  + (A_{1,2}\tau + B_{1,2})(A_{1,2}\tau+B_{1,2})  \\
    &- (A_{1,2}\tau + B_{1,2})^{-1}(A_{1,1}\tau + B_{1,1})  \big(A_{1,2}+(A_{1,2}\tau +B_{1,2})(A_{2,2}\tau) \big).
\end{align*}
\end{lemma}
\begin{lemma}
\label{lemma:tau}
Define the transformation $T$ as \\ $T(f(\tau))$ $=f(\tau)e^{\frac{\int P_1(\tau) d \tau}{2}}$.
Let $y(\tau)=T(g_{w_1}(\tau))$. Then we have
    \begin{align}
    y''(\tau) &=  \big( r_2 \tau^2 + r_1\tau +r_0 +\frac{ r_{-1} }{\tau + A_{1,2}^{-1} B_{1,2}} + \frac{r_{-2} }{(\tau + A_{1,2}^{-1} B_{1,2})^2} \big) y(\tau), \label{eq:yODE}
\end{align}
where 
\begin{align*}
    r_2&= (\frac{1}{4}A_{1,1}^2+A_{1,2}^2+\frac{1}{4}A_{2,2}-\frac{1}{2}A_{1,1}A_{2,2}), \\
    r_1&= \frac{1}{2}A_{1,1}B_{1,1}+2A_{1,2}B_{1,2} -\frac{1}{2}B_{1,1}A_{2,2}-1, \\
    r_0&= \frac{1}{4}B_{1,1}^2+B_{1,2}^2, \\
    r_{-1}&= \frac{1}{2}A_{1,2}^{-1}B_{1,2}( A_{1,1} - A_{2,2} ) -\frac{1}{2}B_{1,1}, \\
    r_{-2}&= -\frac{1}{4}.
\end{align*}
and 
\begin{align}
    &A_{1,1} =   a^2 \sigma(\hat{b})\sigma''(\hat{b}) \overbar{x^2}, \label{eq:A1}  \\
    &B_{1,1} = \frac{1}{\omega} N a  \sigma(\hat{b})\sigma''(\hat{b})  (\overbar{x^2} \overbar{y} - \overbar{x^2y} ),   \\
    &A_{1,2}=A_{2,1} =   a^2  \sigma(\hat{b})\sigma''(\hat{b})  \overbar{x}, \\
    &B_{1,2}=B_{2,1} = \frac{1}{\omega} N a \sigma(\hat{b})\sigma''(\hat{b})  (\overbar{x} \cdot \overbar{y} - \overbar{xy} ),   \\
    &A_{2,2} =    a^2  \sigma(\hat{b})\sigma''(\hat{b}),  \\
    &B_{2,2} = 0. \label{eq:A6}
\end{align}
Note that since $\overbar{x} \ne 0$, we have $A_{1,2} \ne 0$.
\end{lemma}
Hence, by \cref{lemma:epw2,lemma:28}, we need to explore the solvability of $y(\tau)$ in \cref{eq:yODE} to determine the solvability of $\epsilon_{w_1}(t)$. Now, we apply Lemma \ref{lemma:2ndode} to determine the solvability of ODE \eqref{eq:yODE}. By \cref{lemma:2ndode}, if $r_2,r_{-2} \ne 0$ and $-2+\sqrt{1+4r_{-2}}$ is not a nonnegative integer, ODE \cref{eq:yODE} has no Liouvillian solution. Since $-2+\sqrt{1+4 r_{-2}}=-2+\sqrt{1+4 \times (-\frac{1}{4}) } = -2 \not\in \mathbb{N}_{\ge 0}$, we only need to see whether $r_2 \ne 0$. By plugging \eqref{eq:A1}-\eqref{eq:A6} into $r_2$, we have 
\begin{align*}
    &r_2 = a^4 \sigma(\hat{b})^2 \sigma''(\hat{b})^2 (\overbar{x^2})^2 + 4 a^4\sigma(\hat{b})^2 \sigma''(\hat{b})^2 (\overbar{x})^2 + a^2 \sigma(\hat{b}) \sigma''(\hat{b}) - 2 a^4 \sigma(\hat{b})^2 \sigma''(\hat{b})^2 \overbar{x^2}\overbar{x}.
\end{align*}
Hence $r_2 = 0$ if and only if 
\begin{align*}
    & (\overbar{x^2})^2+ 4(\overbar{x})^2-2\overbar{x^2}\overbar{x} + \frac{1}{a^2 \sigma(\hat{b}) \sigma''(\hat{b}) }  \\
    =  & (\overbar{x^2})^2+ 4(\overbar{x})^2-2\overbar{x^2}\overbar{x} + \frac{1}{(\hat{b}_2 - \overbar{y})^2\sigma(\hat{b}) \sigma''(\hat{b})} = 0,
\end{align*}

Since $\hat{b}_2 \in \mathbb{R}$ is arbitrary, we can always choose  $\hat{b}_2 \ne \overbar{y}$ such that $r_2 \ne 0$. 
Therefore, ODE \eqref{eq:yODE} has no Liouvillian solution, and its differential Galois group is $SL_2(\mathbb{C})$. Hence  $\epsilon_{w_1}(t),\epsilon_{b_1}(t)$ also has no Liouvillian solution, and its differential Galois group is also $SL_2(\mathbb{C})$ that is not solvable (also not abelian). Since the differential Galois group of the variational equation is not abelian, the gradient flow is not B-integrable in the meromorphic category by Lemma \ref{lemma:ayoul}. This completes the proof. 
\end{proof}

\section{Absence of the closed-form description of the gradient flow}
In this section, we show demonstrate the unsolvability of the variational equation implies the absence of a closed-form description for the dynamical system. 
\begin{theorem}
\label{theorem:closedform}
    Consider the following $n$-dimensional system
    \begin{align}
        \frac{d \bold{x}(t)}{dt} = F(\bold{x}), \; \bold{x}(t) \in \mathcal{M}, \; t \in \mathbb{C}, \label{eq:ds_th2}
    \end{align}
    where $F(\bold{x})$ is an $n$-dimensional vector-valued function and $\mathcal{M}$ is an $n$-dimensional complex manifold. Let $\gamma(t)$ be an integral flow of system \eqref{eq:ds_th2} with the initial point $\mathbf{p}_1 \in \mathbb{C}^n$. Suppose that the variational equation of $\gamma(t)$ has no Liouvillian solution. Then there is no 
Liouvillain function $\Gamma(\cdot) : \mathbb{C} \times \mathbb{C}^n \to \mathbb{C}^n$ which describes the full dynamics \eqref{eq:ds_th2}. 
\end{theorem}


\begin{proof}
Suppose that such  Liouvillian $\Gamma(\cdot)$ exists.  That is, $\Gamma(t , \mathbf{p})$ is the trajectory with the initial point $\mathbf{p} \in \mathbb{C}$ ( $\Gamma(t,\mathbf{p})$ is an integral curve starting from $\mathbf{p}$ for some $\bold{p} \in \mathbb{C}^n$).
 Let $\eta(t) = \frac{\partial }{\partial \bold{x}}\Gamma(t,\bold{p}_1)\bold{u} \in \mathbb{C}^n$  for some $ \bold{u}=( u_1, ... , u_n) \in \mathbb{C}^n$. Since $\Gamma(\cdot)$ is Liouvillian, $\frac{\partial }{\partial \bold{x}}\Gamma(t,\bold{p}_1)$ is also Liouvillian. 
Since $\Gamma(0,\bold{x})=\bold{x}$, we have $\frac{\partial }{\partial \bold{x}} \Gamma(0,\bold{x})=I_n$ where $\bold{x}  \in \mathbb{C}^n$, and $I_n$ is the $n \times n$ identity matrix. Hence, 
\begin{align*}
    & \eta(0) = \bold{u}, \; \eta'(t) = \frac{\partial}{\partial t} \frac{\partial }{\partial \bold{x}} \Gamma(t,\bold{p}_1)\bold{u} =  \frac{\partial }{\partial \bold{x}} \frac{\partial}{\partial t} \Gamma(t,\bold{p}_1)\bold{u} = \frac{\partial}{\partial \bold{x}} \big( F(\Gamma(t,\bold{p}))  \big) \bold{u} \\
    &= \frac{dF}{d \bold{x}}(\Gamma(t,\bold{p}_1)) \frac{\partial \Gamma}{\partial \bold{x}}(t,\bold{p}_1)  \bold{u}=  \frac{dF}{d \bold{x}}(\Gamma(t,\bold{p}_1))  \eta(t). 
\end{align*}
Therefore, $\eta(t)$ follows the variational equation along $\gamma(t)$ with the initial condition $\eta(0)= \bold{u}$. However, since the variational equation along $\gamma(t)$ has no Liouvillian solution, this contradicts the fact that $\frac{\partial }{\partial \bold{x}}\Gamma(t,\bold{p}_1)$ is Liouvillian. 
\end{proof}
Using Theorem \ref{theorem:main} and Theorem \ref{theorem:closedform}, we can demonstrate that there is no closed-form description of full training dynamics \eqref{eq:gf1}-\eqref{eq:gf4}.  

\begin{corollary}
\label{cor:main}
    There is no Liouvillian function $\Gamma(\cdot) : \mathbb{C} \times \mathbb{C}^4 \to \mathbb{C}^4$ which describes full dynamics \eqref{eq:gf1}-\eqref{eq:gf4}. 
\end{corollary}

\section{Discussion}
We demonstrate the absence of a closed-form description for the gradient flow in two-layer narrow networks. A natural question arises: Is it possible to extend this result to general neural networks? As the number of parameters increases, the dynamics likely become more complex, leading us to expect that the gradient flow in general networks is also sufficiently complex. 
However, providing a strict proof is challenging due to the difficulties associated with determining the differential Galois group for general linear differential equations. For second-order and third-order cases, there are algorithms \cite{kovacic1986algorithm,singer1993liouvillian} to obtain the differential Galois group. Nevertheless, obtaining the differential Galois group for more general differential equations remains very difficult. Despite these challenges, we cautiously conjecture that the differential Galois group is non-abelian for most equations, suggesting the absence of a closed-form description for the gradient flow in general networks. Therefore, a possible future direction would be to explore how well we can approximate the gradient flow to better understand the training dynamics of neural networks.

\section{Conclusion}
We have studied the inherent limitations in comprehending the training dynamics of neural networks.  Our study demonstrates that gradient flow is non-integrable in the meromorphic category by showing that the variational equation of integral curve is not solvable. To establish this, we introduce the differential Galois theory and Morales-Ramis theory. Furthermore, we establish that the unsolvability of the variational equation implies the absence of a closed-form formula to describe the dynamical system.  An important consequence of this finding is the impossibility of representing the training dynamics in basic functions (Liouvillian functions). To the best of our knowledge, this work is the first attempt to investigate the integrability of the gradient flow in neural networks.

While we focus on a simple two-layer narrow network with $\ell^2$ loss, our aim is to generalize our findings to encompass a broader range of neural network architectures, loss functions, and activation functions. Additionally, given that stochastic differential equations are employed to approximate stochastic gradient descent, it is imperative to investigate the predictability on stochastic gradient flow of training dynamics. 

\appendix

\section{Technical lemmas}
In this section, we present several technical lemmas which is used for the proof of main statements. 
\begin{lemma}
\label{lemma:twoODE}
    Consider the following system of ODEs.
    \begin{align}
        \epsilon_1'(t) &= (A_{1,1}t+B_{1,1}) \epsilon_1(t) + (A_{1,2}t+B_{1,2}) \epsilon_2(t) \label{eq:e1p},\\
        \epsilon_2'(t) &= (A_{2,1}t+B_{2,1}) \epsilon_1(t) + (A_{2,2}t+B_{2,2}) \epsilon_2(t), \label{eq:e2p}
    \end{align}
Then we have 
\begin{align*}
    \epsilon_1''(t) = P_1(t) \epsilon_1'(t) + P_2(t) \epsilon_1(t),
\end{align*}
where
\begin{align*}
    P_1(t) &= A_{1,1}t+B_{1,1} + (A_{1,2}t+B_{1,2})^{-1}\big( A_{1,2}+(A_{1,2}t+B_{1,2})(A_{2,2}t+B_{2,2}) \big), \\
    P_2(t) &= \big(A_{1,1} + (A_{1,2}t+B_{1,2})(A_{2,1}t+B_{2,1}) \big), \\
    &- (A_{1,2}t+B_{1,2})^{-1}\big( A_{1,2}+(A_{1,2}t+B_{1,2})(A_{2,2}t+B_{2,2}) \big)  (A_{1,1}t+B_{1,1}).
\end{align*}
\end{lemma}
\begin{proof}
    The proof is presented in \cref{proof:twoODE}.
\end{proof}
\begin{lemma}
\label{lemma:Ntau}
Let
\begin{align*}
    N(\tau) &= -\frac{1}{2}A_{1,2}^{-1}\big(A_{1,2}+(A_{1,2}\tau +B_{1,2})(A_{2,2}\tau) \big)  \\
    & -\frac{1}{2} A_{1,2}^{-2} (A_{1,2}\tau + B_{1,2})( 2(A_{1,2}+A_{2,2})\tau + B_{1,2}A_{2,2} )  \\
    &+\frac{1}{2}A_{1,2}^{-2}(A_{1,1}\tau +B_{1,1})(A_{1,2}\tau + B_{1,2})\big(A_{1,2}+(A_{1,2}\tau +B_{1,2})(A_{2,2}\tau) \big)  \\
    &+\frac{1}{4}A_{1,2}^{-2} \big(A_{1,2}+(A_{1,2}\tau +B_{1,2})(A_{2,2}\tau) \big)^2 \\
    &-A_{1,2}^{-2} (A_{1,2}\tau + B_{1,2})(A_{1,1}\tau + B_{1,1})  \big(A_{1,2}+(A_{1,2}\tau +B_{1,2})(A_{2,2}\tau) \big). \nonumber
\end{align*}
Then we have 
\begin{align*}
N(\tau) =  N_2(\tau)(\tau + A_{1,2}^{-1} B_{1,2})^2 + N_1(\tau) (\tau + A_{1,2}^{-1} B_{1,2}) + N_0(\tau),
\end{align*}
where
\begin{align*}
    N_2(\tau) &=(\frac{1}{4}A_{2,2}-\frac{1}{2}A_{1,1}A_{2,2} )\tau^2 -(\frac{1}{2}B_{1,1}A_{2,2}+1)\tau -\frac{1}{2}A_{1,1},   \\
    N_1(\tau) &=  \frac{1}{2}A_{1,2}^{-1}B_{1,2}( A_{1,1} - A_{2,2} ) -\frac{1}{2}B_{1,1},  \\
    N_0(\tau) &= -\frac{1}{4}.
\end{align*}
\end{lemma}
\begin{proof}
The proof is presented in \cref{proof:Ntau}.
\end{proof}
\section{Proofs of lemmas}
\label{sec:proof_lemmas}
\subsection{Proof of \cref{lemma:icurve}}
By substituting curve \eqref{eq:ic} into \eqref{eq:gf1}-\eqref{eq:gf4}, we have 
\begin{small}
\begin{align*}
     w_1'(t)&=  -\sum_{i=1}^N ( a \sigma(\hat{b}) e^{-\omega t} \sigma( \hat{b} ) + a e^{- \omega t} + \overbar{y} -y_i) (a \sigma(\hat{b}) e^{-\omega t} \sigma'(\hat{b}) ) x_i  = 0, \\ 
     b_1'(t)&= -\sum_{i=1}^N ( a \sigma(\hat{b}) e^{-\omega t} \sigma( \hat{b} ) + a e^{- \omega t} + \overbar{y} -y_i) (a \sigma(\hat{b}) e^{-\omega t} \sigma'(\hat{b}) )   = 0,\\ 
     w_2'(t)& = -\sum_{i=1}^N ( a \sigma(\hat{b}) e^{-\omega t} \sigma( \hat{b} ) + a e^{- \omega t} + \overbar{y} -y_i)  \sigma(\hat{b})  \\
 & =- \sum_{i=1}^N a\sigma(\hat{b}) ( \sigma(\hat{b})^2 + 1 )   e^{- \omega t}  + (\overbar{y}-y_i) \sigma(\hat{b}) = -  a \sigma(\hat{b}) \omega e^{- \omega t} =  \frac{d}{dt}(a \sigma(\hat{b}) e^{-\omega t}),  \\
     b_2'(t)&=  -\sum_{i=1}^N ( a \sigma(\hat{b}) e^{-\omega t} \sigma( \hat{b} ) + a e^{- \omega t} + \overbar{y} -y_i)   \\
     & =- \sum_{i=1}^N a ( \sigma(\hat{b})^2 + 1 )   e^{- \omega t}  + (\overbar{y}-y_i)  = -  a \omega e^{- \omega t} =  \frac{d}{dt}(a e^{-\omega t}+\overbar{y}).
\end{align*}
\end{small}
This completes the proof.
\newpage 
\subsection{Proof of \cref{lem:vari}}
We calculate every element of the $4 \times 4$ matrix in \eqref{eq:ve_matrix}.
\begin{small}
\begin{align*}
    \frac{\partial w_1'(t) } {\partial w_1} &= -\sum_{i=1}^N  \big( w_2^2 (\sigma'(w_1 x_i+ b_1))^2  +   (w_2\sigma(w_1 x_i+ b_1) + b_2 -y_i) (w_2\sigma''(w_1 x_i + b_1) ) \big) x_i^2, \\
    \frac{\partial w_1'(t) } {\partial b_1} &= \frac{\partial b_1'(t) } {\partial w_1} \\
    &=  -\sum_{i=1}^N  \big( w_2^2 (\sigma'(w_1 x_i+ b_1))^2  +   (w_2\sigma(w_1 x_i+ b_1) + b_2 -y_i) (w_2\sigma''(w_1 x_i + b_1) ) \big) x_i, \\
    \frac{\partial w_1'(t) } {\partial w_2} &= \frac{\partial w_2'(t) } {\partial w_1} =  -\sum_{i=1}^N  \big(  2w_2\sigma(w_1 x_i+ b_1) + b_2 -y_i  \big) \sigma'(w_1 x_i + b_1) x_i, \\
    \frac{\partial w_1'(t) } {\partial b_2} &= \frac{\partial b_2'(t) } {\partial w_1} =  -\sum_{i=1}^N w_2 \sigma'(w_1 x_i + b_1)  x_i, \\
    \frac{\partial b_1'(t) } {\partial b_1} &= -\sum_{i=1}^N  \big( w_2^2 (\sigma'(w_1 x_i+ b_1))^2  +   (w_2\sigma(w_1 x_i+ b_1) + b_2 -y_i) (w_2\sigma''(w_1 x_i + b_1) ) \big), \\
    \frac{\partial b_1'(t) } {\partial w_2} &= \frac{\partial w_2'(t) } {\partial b_1} =    -\sum_{i=1}^N  \big(  2w_2\sigma(w_1 x_i+ b_1) + b_2 -y_i  \big) \sigma'(w_1 x_i + b_1)  , \\
    \frac{\partial b_1'(t) } {\partial b_2} &= \frac{\partial b_2'(t) } {\partial b_1} =  -\sum_{i=1}^N w_2 \sigma'(w_1 x_i + b_1), \\
    \frac{\partial w_2'(t) } {\partial w_2} &= -\sum_{i=1}^N (\sigma(w_1 x_i+ b_1) )^2 , \\
    \frac{\partial w_2'(t) } {\partial b_2} &= \frac{\partial b_2'(t) } {\partial w_2} =  -\sum_{i=1}^N \sigma(w_1 x_i+ b_1), \\
    \frac{\partial b_2'(t) } {\partial b_2} &= -\sum_{i=1}^N 1. 
\end{align*}
\end{small}
\newpage 

By substituting curve \eqref{eq:ic} into above equations, we have
\begin{align*}
    &\frac{\partial w_1'(t) } {\partial w_1} = -\sum_{i=1}^N  \big( ( a \sigma(\hat{b})^2 e^{-\omega t} + a e^{-\omega t} + \overbar{y} -y_i) ( a \sigma(\hat{b}) e^{- \omega t }\sigma''(\hat{b}) ) \big) x_i^2 \\
    &= -\sum_{i=1}^N  \big( (  a \frac{\omega}{N}  e^{-\omega t} + \overbar{y} -y_i) ( a \sigma(\hat{b}) e^{- \omega t }\sigma''(\hat{b}) ) \big) x_i^2,  \\
    &\frac{\partial w_1'(t) } {\partial b_1} = \frac{\partial b_1'(t) } {\partial w_1} = -\sum_{i=1}^N  \big( (  a \frac{\omega}{N}  e^{-\omega t} + \overbar{y} -y_i) ( a \sigma(\hat{b}) e^{- \omega t }\sigma''(\hat{b}) ) \big) x_i, \\
    &\frac{\partial w_1'(t) } {\partial w_2} = \frac{\partial w_2'(t) } {\partial w_1} = \frac{\partial w_1'(t) } {\partial b_2} = \frac{\partial b_2'(t) } {\partial w_1} = 0, \\
    & \frac{\partial b_1'(t) } {\partial b_1} = -\sum_{i=1}^N  \big( (  a \frac{\omega}{N}  e^{-\omega t} + \overbar{y} -y_i) ( a \sigma(\hat{b}) e^{- \omega t }\sigma''(\hat{b}) ) \big), \\
    & \frac{\partial b_1'(t) } {\partial w_2} = \frac{\partial w_2'(t) } {\partial b_1} = \frac{\partial b_1'(t) } {\partial b_2} = \frac{\partial b_2'(t) } {\partial b_1}= 0, \\
    & \frac{\partial w_2'(t) } {\partial w_2} = -\sum_{i=1}^N (\sigma(\hat{b}) )^2,  \frac{\partial w_2'(t) } {\partial b_2} = \frac{\partial b_2'(t) } {\partial w_2}= -\sum_{i=1}^N \sigma(\hat{b}), \frac{\partial b_2'(t) } {\partial b_2} = -\sum_{i=1}^N 1.
\end{align*}
Therefore, the variational equation along the integral curve $\gamma(t)$ has the form of \cref{eq:ve1,eq:ve2,eq:ve3,eq:ve4}.

\subsection{Proof of Lemma \ref{lemma:epw2}}
\label{proof:epw2}
By \cref{eq:ve3,eq:ve4}, we have 
\begin{align*}
    \epsilon_{w_2}'(t) &=  - N \sigma(\hat{b}_1)^2 \epsilon_{w_2}(t)  - N \sigma(\hat{b}_1) \epsilon_{b_2}(t), \\
    \epsilon_{b_2}'(t) &=  - N \sigma(\hat{b}_1) \epsilon_{w_2}(t)  -N  \epsilon_{b_2}(t).
\end{align*}
By Lemma \ref{lemma:twoODE}, we have
\begin{align*}
    \epsilon_{w_2}''(t) &= (- N \sigma(\hat{b}_1)^2-N)  \epsilon_{w_2}'(t) = 0.
\end{align*}
Therefore we can reduce above ODE to the first order ODE, and its corresponding Galois group is $\mathbb{C}^*$. Because $\epsilon_{w_2}'(t) = \sigma(\hat{b}) \epsilon_{b_2}'(t)$, the differential Galois group for $\epsilon_{b_2}'(t)$ is same as $\epsilon_{w_2}'(t)$.  

\subsection{Proof of \cref{lemma:28}}
By \cref{eq:ve1,eq:ve2}, we have 
\begin{align*}
    &\epsilon_{w_1}'(t) = \\
    &(-  a^2 \omega \sigma(\hat{b})\sigma''(\hat{b}) e^{-2\omega t}  \overbar{x^2} - N a \sigma(\hat{b})\sigma''(\hat{b}) e^{-\omega t}  \overbar{x^2} \overbar{y} + N a\sigma(\hat{b})\sigma''(\hat{b}) e^{-\omega t} \overbar{x^2y} ) \epsilon_{w_1}(t) \\
    &+ (-  a^2 \omega \sigma(\hat{b})\sigma''(\hat{b}) e^{-2\omega t}  \overbar{x} - N a \sigma(\hat{b})\sigma''(\hat{b}) e^{-\omega t}  \overbar{x} \cdot  \overbar{y} + N a\sigma(\hat{b})\sigma''(\hat{b}) e^{-\omega t} \overbar{xy} ) \epsilon_{b_1}(t),  \\
    &\epsilon_{b_1}'(t) = \\
    & (-  a^2 \omega \sigma(\hat{b})\sigma''(\hat{b}) e^{-2\omega t}  \overbar{x} - N a \sigma(\hat{b})\sigma''(\hat{b}) e^{-\omega t}  \overbar{x} \cdot \overbar{y} + N a\sigma(\hat{b})\sigma''(\hat{b}) e^{-\omega t} \overbar{xy} ) \epsilon_{w_1}(t), \\
    &+ (-  a^2 \omega \sigma(\hat{b})\sigma''(\hat{b}) e^{-2\omega t}  - N a \sigma(\hat{b})\sigma''(\hat{b}) e^{-\omega t}  \overbar{y} + N a\sigma(\hat{b})\sigma''(\hat{b}) e^{-\omega t} \overbar{y} ) \epsilon_{b_1}(t).
\end{align*}
Let $g_{w_1}(\tau) = \epsilon_{w_1}'(t)$ and $g_{b_1}(\tau) = \epsilon_{b_1}'(t)$.  Since $\frac{d f}{d \tau} = \frac{d f}{d t} \frac{dt}{d \tau} = -\frac{d f}{d t}   \frac{1}{\omega \tau} $ for $f=f(\tau)$, we have 
\begin{align*}
    g_{w_1}'(\tau) = ( A_{1,1}\tau +B_{1,1} )g_{w_1}(\tau) + ( A_{1,2}\tau +B_{1,2} )g_{b_1}(\tau), \\
    g_{b_1}'(\tau) = ( A_{2,1}\tau +B_{2,1} )g_{w_1}(\tau) + ( A_{2,2}\tau +B_{2,2} )g_{b_1}(\tau), \\
\end{align*}
where
\begin{align*}
    &A_{1,1} =   a^2 \sigma(\hat{b})\sigma''(\hat{b}) \overbar{x^2}, \\
    &B_{1,1} = \frac{1}{\omega} N a  \sigma(\hat{b})\sigma''(\hat{b})  (\overbar{x^2} \overbar{y} - \overbar{x^2y} ),  \\
    &A_{1,2}=A_{2,1} =   a^2  \sigma(\hat{b})\sigma''(\hat{b})  \overbar{x}, \\
    &B_{1,2}=B_{2,1} = \frac{1}{\omega} N a \sigma(\hat{b})\sigma''(\hat{b})  (\overbar{x} \cdot \overbar{y} - \overbar{xy} ),   \\
    &A_{2,2} =    a^2  \sigma(\hat{b})\sigma''(\hat{b}),   \\
    &B_{2,2} = 0.
\end{align*}
By Lemma \ref{lemma:twoODE}, we have the following second-order ODE for $g_{w_1}(\tau)$ 
\begin{align*}
    &g_{w_1}''(\tau) = P_1(\tau) g_{w_1}'(\tau) + P_2(\tau)g_{w_1}(\tau),
\end{align*}
where
\begin{align*}
    & P_1(\tau) =  A_{1,1}\tau +B_{1,1}+ (A_{1,2}\tau + B_{1,2})^{-1} \big(A_{1,2}+(A_{1,2}\tau +B_{1,2})(A_{2,2}\tau) \big),  \\
    &P_2(\tau) = A_{1,1}  + (A_{1,2}\tau + B_{1,2})(A_{1,2}\tau+B_{1,2})  \\
    &- (A_{1,2}\tau + B_{1,2})^{-1}(A_{1,1}\tau + B_{1,1})  \big(A_{1,2}+(A_{1,2}\tau +B_{1,2})(A_{2,2}\tau) \big).
\end{align*}
\subsection{Proof of \cref{lemma:tau}}

By taking the transformation \\
$T(f(\tau))=f(\tau)e^{\frac{\int P_1(\tau) d \tau}{2}}$ to ODE \eqref{eq:28}, we can have the simplified equation for $y(\tau)=T(\epsilon_{w_1}(\tau))$: 
\begin{align}
    y''(\tau) &=  (-\frac{P_1'(\tau)}{2}+\frac{P_1(\tau)^2}{4}+P_2(\tau))y(\tau) \nonumber \\
    &= -\frac{A_{1,1}}{2} + \frac{(A_{1,1}\tau +B_{1,1})^2}{4} + A_{1,1}  + (A_{1,2}\tau + B_{1,2})^2 +\frac{ N(\tau)  }{( \tau + A_{1,2}^{-1} B_{1,2})^2}, \label{eq:veode}
\end{align}
where 
\begin{align*}
     N(\tau) = &-\frac{1}{2}A_{1,2}^{-1}\big(A_{1,2}+(A_{1,2}\tau +B_{1,2})(A_{2,2}\tau) \big)  \\
     &-\frac{1}{2} A_{1,2}^{-2} (A_{1,2}\tau + B_{1,2})( 2(A_{1,2}+A_{2,2})\tau + B_{1,2}A_{2,2} )  \\
    &+\frac{1}{2}A_{1,2}^{-2}(A_{1,1}\tau +B_{1,1})(A_{1,2}\tau + B_{1,2})\big(A_{1,2}+(A_{1,2}\tau +B_{1,2})(A_{2,2}\tau) \big) \\ 
    &+ \frac{1}{4}A_{1,2}^{-2} \big(A_{1,2}+(A_{1,2}\tau +B_{1,2})(A_{2,2}\tau) \big)^2 \\
    &-A_{1,2}^{-2} (A_{1,2}\tau + B_{1,2})(A_{1,1}\tau + B_{1,1})  \big(A_{1,2}+(A_{1,2}\tau +B_{1,2})(A_{2,2}\tau) \big). \nonumber
\end{align*}
By organizing \cref{eq:veode}, we have
\begin{align*}
    y''(\tau) &=  \big( r_2 \tau^2 + r_1\tau +r_0 +\frac{ r_{-1} }{\tau + A_{1,2}^{-1} B_{1,2}} + \frac{r_{-2} }{(\tau + A_{1,2}^{-1} B_{1,2})^2} \big) y(\tau), 
\end{align*}
where 
\begin{align*}
    r_2&= (\frac{1}{4}A_{1,1}^2+A_{1,2}^2+\frac{1}{4}A_{2,2}-\frac{1}{2}A_{1,1}A_{2,2}), \\
    r_1&= \frac{1}{2}A_{1,1}B_{1,1}+2A_{1,2}B_{1,2} -\frac{1}{2}B_{1,1}A_{2,2}-1, \\
    r_0&= \frac{1}{4}B_{1,1}^2+B_{1,2}^2, \\
    r_{-1}&= \frac{1}{2}A_{1,2}^{-1}B_{1,2}( A_{1,1} - A_{2,2} ) -\frac{1}{2}B_{1,1}, \\
    r_{-2}&= -\frac{1}{4}.
\end{align*}

\section{Differential Galois Theory}
\label{section:differentialGalois}
In this section, we provide a brief introduction to the differential Galois theory. For more information, please refer to \cite{churchill2006liouville,hubbard2011first,van2012galois}.

\subsection{Classical Galois theory}
We first present some essentials of classical Galois theory. Classical Galois theory deals with the representation of solutions to polynomial equations by a finite number of additions, multiplications, and $n$-th roots of rational numbers.

\begin{definition}[field extension]
    Let $L$ be a field and $K$ be a subfield of $K$. $K \subset L$ is called a field extension. The larger field $L$ is a $K$-vector space. The degree of a field extension $K \subset L$ is the dimension of the vector space, \textit{i.e.}, 
    \begin{equation*}
        [L:K] = \dim_K L.
    \end{equation*}
    $\alpha$ is algebraic if it is a root of a non-zero polynomial with coefficients in $K$. If every element of $L$ is algebraic over $K$, then the extension $K \subset L$ is called an algebraic extension. If $\alpha$ is not a root of any polynomial with coefficients in $K$, $\alpha$ is transcendental. An extension $K \subset L$ is a transcendental extension if $L$ has a transcendental element over $K$. 
\end{definition}

\begin{proposition}
    For an algebraic extension $K \subset K(\alpha)$, the extension degree $[K(\alpha) : K]$ equals the degree of the minimal polynomial $p(x)$ such that $p(\alpha) = 0$. If the extension $K \subset K(\alpha)$ is transcendental, the extension degree is infinite.
\end{proposition}

\begin{definition}[Galois group]
    Let $K \subset L$ be a algebraic field extension. The extension $K \subset L$ is called a normal extension if every irreducible polynomial over $K$ that has a root in $L$ splits into linear factors in $L$. The extension $K \subset L$ is called a separable extension if for every $\alpha \in L$, the minimal polynomial of $\alpha$ has no repeated roots. The extension $K \subset L$ is called a Galois extension if it is both normal and separable. If the extension $K \subset L$ is Galois, then its corresponding Galois group $Gal(L/K)$ is defined as the group of field automorphisms of $L$ that fix $K$. 
\end{definition}

\begin{proposition}
   Let $K \subset L = K(\alpha_1,\alpha_2, ... ,\alpha_n)$ be a Galois extension, where $\alpha_1,\alpha_2, ... ,\alpha_n$ are roots of an irreducible polynomial $p(x)$ of degree $n$. Then its corresponding Galois group $Gal(L/K)$ is a subgroup of the symmetric group $S_n$. 
\end{proposition}

\begin{definition}[radical extension, solvable by radicals]
    Let $K$ be a field. $L = K(l)$ is called a radical extension of $K$ if $l^n=k$ for some $k \in K , n \in \mathbb{N}$. \\
    Let $K=\mathbb{Q}$ be the base field. For the field extension $K \subset K(\alpha)$,  if there exists a finite sequence of intermediate field extensions 
    \begin{align*}
        K \subset K_1 \subset ... \subset K_m = K(\alpha),
    \end{align*}
such that $ K_{i-1} \subset K_i$ is radical, $\alpha$ is called solvable by radicals.
\end{definition}
If $\alpha$ is solvable by radicals, $\alpha$ is representable by a finite number of additions, multiplications, and $n$-th roots of rational numbers.  

\begin{definition}[solvable group]
    Let $G$ be a group. $G$ is called solvable if there exists a subnormal series 
    \begin{align*}
        1 = G_0  \triangleleft G_1 ... \triangleleft G_m = G,
    \end{align*}
such that $G_{i-1}$ is a normal subgroup of $G_{i}$ and the quotient group $G_i/G_{i-1}$ is abelian for $i=1,...,m-1$. 
\end{definition}
\begin{lemma}
Let $K = \mathbb{Q}$ be the base field. 
    Let $\alpha_1, ... , \alpha_n$ be $n$ solutions of $n$-th order polynomial 
    \begin{align*}
        a_n x^n + a_{n-1}x^{n-1}+ ... + a_1x + a_0 = 0, 
    \end{align*}
where $a_n \ne 0$, $a_n , ... , a_0 \in \mathbb{Q}$. 
Then $\alpha_1,...,\alpha_n$ are solvable by radicals if \\
$Gal(K(\alpha_1, ... , \alpha_n)/K)$ is solvable. 
\end{lemma}

\subsection{Differential Galois theory}
Next, we present some essentials of the differential Galois theory. Differential Galois theory deals with  the representation of solutions of linear ODEs using a finite number of operations including additions, multiplications, $n$-th roots, exponentials, and anti-derivatives of rational functions.

\begin{definition}[differential field]
    Let $K$ be a field. An additive group homomorphism $ (') : K \to K$ is a derivation, if the Leibniz rule
    \begin{equation*}
        (k_1 k_2)' = k_1'k_2 + k_1k_2',
    \end{equation*}
    holds for all $k_1,k_2 \in K$. \\ 
    $K$ is called a differential field if it is equipped with the derivation. \\ 
    The subfield $Con(K)$ is called the constants of $K$ if
    \begin{equation*}
        Con(K) = \{ k \in K : k' =0 \}.
    \end{equation*}
\end{definition}

\begin{definition}[exponential extension, primitive extension]
    Let $K$ be a differential field. $L=K(l)$ is called an exponential extension of $K$ if $l$ is transcendental over $K$ and  
    \begin{equation*}
        \frac{l'}{l} = k',
    \end{equation*}
    for some $k \in K$. \\
Similarly, $L = K(l)$ is called a primitive extension of $K$ if $l$ is transcendental over $K$ and 
    \begin{equation*}
        l' = k  \in K,
    \end{equation*}
for some $k \in K$. 
This is analogues to the logarithms and exponentials, where $l=e^k$ and $l = \int k$, respectively. 
\end{definition}
Note that if $l= \log(k)$ is transcendental, then $l$ is primitive since $\log(k) = \int \frac{k'}{k}$. 
\begin{definition}[differential Galois group]
\label{def:differentialGalois}
    Let $K \subset L$ be a differential field extension. Its corresponding differential Galois group $G := Gal(L/K)$ is defined as the group of differential field automorphisms of $L$ that fix $K$, and satisfies
    \begin{equation*}
        g(l') = g(l)',
    \end{equation*}
    for all $g \in Gal(L/K)$ and $l \in L$. 
\end{definition}
\begin{proposition}
     Let $K \subset L$ be a differential field extension of degree $n$ linear ODE 
     \[ L(y) = a_n y^{(n)} + a_{n-1}y^{(n-1)} + ... + a_1 y + a_0=0.  \]
    Then its corresponding differential Galois group $Gal(L/K)$ is a subgroup of a general linear group $GL_n(\mathbb{C})$. 
\end{proposition}
\begin{definition}[Liouvillian extension, Liouvillian function]
    Let $K \subset L$ be a differential field extension. We say $l \in L$ is Liouvillian over $K$ if $l$ is either algebraic, primitive, or exponential over $K$. Similarly, a differential field extension $K \subset L$ is Liouvillian if there exists a finite sequence of
intermediate differential field extensions 
\begin{equation}
    K = K_0 \subset K_1 \subset ... \subset K_n = L,
\end{equation}
such that $K_{i+1} = K_i(l_i)$ and $l_i$ is Liouvillian over $K_i \; $ for $1 \le i \le n$. 
$l$ is called Liouvillian over $K$ if $K \subset K(l)$ is a Liouvillian extension.  \\
If $l$ is Liouvillian over $\mathbb{C}(x)$, we simply call $l$ be a Liouvillian function. 
\end{definition}
If $l$ is Liouvillian, $l$ is representable by a finite number of additions, multiplications, $n$-th roots, exponentials, and anti-derivatives of algebraic functions.  
\begin{lemma}
     Let $\alpha$ be a solution of degree $n$ linear ODE 
     \[ L(y) = a_n y^{(n)} + a_{n_1}y^{(n-1)} + ... + a_1 y + a_0=0,  \]
     and $ K \subset L $ be a differential field extension of $L(y)$. 
     $\alpha$ is Liouvillian if the identity component $Gal^0(L/K)$ of differential Galois group $Gal(L/K)$ is solvable.
\end{lemma}

\begin{remark}
    $SL_n(\mathbb{C})$, $GL_n(\mathbb{C})$ is not solvable if $n \ge 2$. Therefore, in general, solutions of second order linear ODEs are not representable (\textit{cf.} Bessel equations).
\end{remark}

\begin{lemma}
    For $n$-th order linear ODE 
    \[L(y) = a_n y^{(n)} + a_{n-2}y^{(n-2)} + ... + a_1 y + a_0=0,  \] 
    its differential Galois group is an unimodular group (\textit{i.e.}, $Gal(L(y)) \subset SL_n(\mathbb{C}) )$. 
\end{lemma}



\subsection{Differential Galois group of second-order ODE}
In this section, we present the differential Galois group and the solvability of second-order ODE.

\begin{lemma}[Kovacic's algoritm \cite{kovacic1986algorithm}]
\label{lemma:kovacic}
Consider the following second order ODE
\begin{align}
     L(y) =  y''(t)- ry(t) = 0. \label{eq:2ndODE}
\end{align}
    \begin{itemize}
        \item Case 1 : Every pole of $r$ has an even order, or else it has an order of $1$. The order of $r$ at $\infty$ is even or  greater than $2$. In addition, $r$ satisfies Condition \ref{condition:case1}.
        \item Case 2: $r$ has at least one pole with an order greater that $2$ or an order of $2$.  In addition, $r$ satisfies Condition \ref{condition:case2}. 
        \item Case 3: $r$ has a pole with an order of $1$ or $2$. The order of $r$ at $\infty$ is at least $2$.  In addition, $r$ satisfies Condition \ref{condition:case3}.
    \end{itemize}
If none of the above cases holds, then the differential Galois group  $DGal(L(y))$ is $SL_2(\mathbb{C})$. Therefore, the differential equation \eqref{eq:2ndODE} has no Liouvillian solution. 
\end{lemma}

\begin{customcondition}{1}[Condition for Case 1]
\label{condition:case1}
    Let $\Gamma$ be the set of poles of $r$. For each $c \in \Gamma$, we define a rational function $[\sqrt{r}]_c$ and $\alpha_c^+, \alpha_c^- \in \mathbb{C}$ as described below. 
\begin{itemize}
    \item If $c$ is a pole of order $1$, then
    \begin{align*}
        [\sqrt{r}]_c = 0, \; \alpha_c^+=\alpha_c^-=1.
    \end{align*}
    \item If $c$ is a pole of order $2$, then $b_c$ is the coefficient of $\frac{1}{(x-c)^{2}}$ in the partial fraction expansion of $r$ and 
    \begin{align*}
        [\sqrt{r}]_c = 0, \; \alpha_c^+= \frac{1}{2}( 1 + \sqrt{1+4b_c}), \; \alpha_c^-= \frac{1}{2}( 1 - \sqrt{1+4b_c}).
    \end{align*}
    \item If $c$ is a pole of order $2v_c \ge 4$, then $[\sqrt{r}]_c$ is the sum of terms involving $\frac{1}{(x-c)^i}$ for $2 \le i \le v_c$ in the Laurent series expansion of $\sqrt{r}$ (not $r$) at $c$. Let $a_c$ be a coefficient of $\frac{1}{(x-c)^{v_c}}$ in $[\sqrt{r}]_c$, and  $b_c$ be a coefficient of $\frac{1}{(x-c)^{v_c+1}}$ in $r$ minus a coefficient of $\frac{1}{(x-c)^{v_c+1}}$ in Laurent series expansion of $\sqrt{r}$ at $c$. Then 
    \begin{align*}
       \alpha_c^+ = \frac{1}{2}(\frac{b_c}{a_c} + v_c), \;  \alpha_c^- = \frac{1}{2}(-\frac{b_c}{a_c} + v_c),
    \end{align*}
\end{itemize}
We define a rational function $[\sqrt{r}]_\infty$ and $\alpha_\infty^+, \alpha_\infty^- \in \mathbb{C}$ as described below.
\begin{itemize}
    \item The order of $r$ at $\infty$ is $>2$, then 
    \begin{align*}
        [\sqrt{r}]_\infty = 0, \; \alpha_\infty^+= 0, \; \alpha_\infty^-= 1.
    \end{align*}
    \item The order of $r$ at $\infty$ is $2$, then $b_\infty$ is the coefficient of $\frac{1}{x^2}$ in the Laurent series expansion of $r$ at $\infty$ and 
    \begin{align*}
        [\sqrt{r}]_\infty = 0, \; \alpha_\infty^+= \frac{1}{2}(1+\sqrt{1+4b_\infty}) , \; \alpha_\infty^-= \frac{1}{2}(1-\sqrt{1+4b_\infty}).
    \end{align*}
    \item The order of $r$ at $\infty$ is $ -2 v_\infty \le 0$, then $[\sqrt{r}]_\infty$ is the sum of terms involving $x^i$ for $0 \le i \le v_\infty$ in the Laurent series expansion of $\sqrt{r}$ (not $r$) at $\infty$. Let $a_\infty$ be a coefficient of $x^{v_\infty}$ in $[\sqrt{r}]_\infty$, and  $b_\infty$ be a coefficient of $x^{v_\infty-1}$ in $r$ minus a coefficient of $x^{v_\infty-1}$  in $([\sqrt{r}]_\infty)^2$.
    Then 
    \begin{align*}
       \alpha_\infty^+ = \frac{1}{2}(\frac{b_\infty}{a_\infty} - v_\infty), \;  \alpha_\infty^- = \frac{1}{2}(-\frac{b_\infty}{a_\infty} - v_\infty),
    \end{align*}
\end{itemize}
Then the condition is as follows. For any $s_1,s_2 \in \{ +, - \}$, 
\begin{align*}
    \alpha_\infty^{s_1} - \sum_{c \in \Gamma} \alpha_c^{s_2} \not\in \mathbb{N}_{\ge 0}.
\end{align*}
\end{customcondition}

\begin{customcondition}{2}[Condition for Case 2]
\label{condition:case2}
    Let $\Gamma$ be the set of poles of $r$. For each $c \in \Gamma$, we define a set $E_c$ as described below. 
\begin{itemize}
    \item If $c$ is a pole of order $1$, then $E_c = \{ 4\}$.
    \item If $c$ is a pole of order $2$, then $b_c$ is the coefficient of $\frac{1}{(x-c)^{2}}$ in the partial fraction expansion of $r$ and 
    \begin{align*}
        E_c = \{ 2, 2+2\sqrt{1+4 b_c},2-2\sqrt{1+4 b_c}  \} \cap \mathbb{Z}.
    \end{align*}
    \item If $c$ is a pole of order $v_c > 2$, then $E_c = \{ v_c\}$.
\end{itemize}
We define the set $E_\infty$ as described below.
\begin{itemize}
    \item The order of $r$ at $\infty$ is $>2$, then $E_\infty  = \{ 0,2,4\}$. 
    \item The order of $r$ at $\infty$ is $2$,
    then $b_\infty$ is the coefficient of $\frac{1}{x^2}$ in the Laurent series expansion of $r$ at $\infty$ and 
    \begin{align*}
        E_\infty = \{ 2, 2+2\sqrt{1+4 b_\infty},2-2\sqrt{1+4 b_\infty}  \} \cap \mathbb{Z}.
    \end{align*}
    then $E_\infty  = \{ 0,2,4\}$. 
    \item The order of $r$ at $\infty$ is $v_\infty <2$, then $E_\infty = \{ v_\infty \}$. 
\end{itemize}
Then the condition is as follows. For any $e_\infty \in E_\infty , e_c \in E_c$, 
\begin{align*}
    \frac{1}{2}( e_\infty - \sum_{c \in \Gamma} e_c)  \not\in \mathbb{N}_{\ge 0}.
\end{align*}
\end{customcondition}

\begin{customcondition}{3}[Condition for Case 3]
\label{condition:case3}
    Let $\Gamma$ be the set of poles of $r$. For each $c \in \Gamma$, we define the set $E_c$ as described below. 
\begin{itemize}
    \item If $c$ is a pole of order $1$, then $E_c = \{ 12 \}$.
    \item If $c$ is a pole of order $2$, then $b_c$ is the coefficient of $\frac{1}{(x-c)^{2}}$ in the partial fraction expansion of $r$ and 
    \begin{align*}
        E_c = \{ 6+ \frac{12k}{n}\sqrt{1+4b_c} : k=0,\pm 1, \pm 2, ... , \pm \frac{n}{2}, \; n \in \{4,6,12 \}  \} \cap \mathbb{Z}.
    \end{align*}
\end{itemize}
We define the set $E_\infty$ as described below. If the Laurent series expansion of $r$ at $\infty$ is 
\begin{align*}
    r = \gamma x^{-2} + ... \; ( \gamma \in \mathbb{C}), 
\end{align*}
then 
\begin{align*}
    E_\infty = \{ 6 + \frac{12k}{n} \sqrt{1+4\gamma} : k=0, \pm 1 , \pm 2 , ... , \pm \frac{n}{2}, \; n \in \{4,6,12\} \} \cup \mathbb{Z}. 
\end{align*}
Then the condition is as follows. For any $e_\infty \in E_\infty , e_c \in E_c$, 
\begin{align*}
    \frac{1}{12}( e_\infty - \sum_{c \in \Gamma} e_c)  \not\in \mathbb{N}_{\ge 0}.
\end{align*}
\end{customcondition}

\section{Proofs of technical lemmas}
\subsection{Proof of \cref{lemma:2ndode}}
\label{proof:2ndode}
\begin{lemma*}{1}
    Consider the following second order ODE
\begin{align*}
    L(y) &= y(t)'' - r y(t), \\
    r &= a_2 t^2 + a_1 t + a_0 + \frac{a_{-1}}{t+d}+ \frac{b}{(t+d)^2},
\end{align*}
where $a_2,a_1,a_0,a_{-1},b,d \in \mathbb{C}$. If $a_2, b\ne 0$ and  $-2+\sqrt{1+4b}$ is not a nonnegative integer,  then $L(y)$ has no Liouvillian solution and its differential Galois group is $Gal(L(y)) = SL_2(\mathbb{C})$. 
\end{lemma*}
\begin{proof}
In this proof, we actively use \cref{lemma:kovacic} (Kovacic's algorithm).
Since $a_2,a_2,b \ne 0$, $r$ has a pole at $t=0$ of order 2 and the order of $r$ at $\infty$ is $-2$, Case 1 and 2 in Lemma \ref{lemma:kovacic} are possible. 

\textbf{Case 1}:
The coefficient of a pole at $t=0$ is $b_0 = b$. 
\begin{align*}
    \alpha_0^+ = \frac{1}{2}(1+\sqrt{1+4b}), \; \alpha_0^- = \frac{1}{2}(1-\sqrt{1+4b})
\end{align*}
The order of $r$ at $\infty$ is $-2$. We have $v_\infty = 1$, 
\begin{align*}
    [\sqrt{r}]_\infty =   b_\infty = 0 , \; \alpha_\infty^+ = \alpha_\infty^-= -\frac{1}{2}
\end{align*}
Therefore if $\sqrt{1+4b}$ is not an even integer,  for any $s_1,s_2 \in \{ +, - \}$, we have
\begin{align*}
    \alpha_\infty^{s_1} - \sum_{c \in \Gamma} \alpha_c^{s_2} = -\frac{1}{2} -  \frac{1}{2}(1 \pm \sqrt{1 + 4b}) = -1 \mp \frac{\sqrt{1 + 4b}}{2}  \not\in \mathbb{N}_{\ge 0}.
\end{align*}

\textbf{Case 2}:
Since $t=0$ is a pole of order 2, we have
\begin{align*}
    E_0 = \{ 2 , 2+2\sqrt{1+4b}, 2-2\sqrt{1+4b} \} \cap \mathbb{Z}
\end{align*}
Since the order of $r$ at $\infty$ is $-2$, we have
\begin{align*}
    E_\infty = \{ -2 \}. 
\end{align*}
Therefore if $\sqrt{1+4b}$ is not an integer, for any $e_\infty \in E_\infty , e_c \in E_c$ we have
\begin{align*}
    \frac{1}{2}( e_\infty - \sum_{c \in \Gamma} e_c)  = -2 , -2 \mp \sqrt{1+4b} \not\in \mathbb{N}_{\ge 0}.
\end{align*}
\end{proof}

\subsection{Proof of \cref{lemma:twoODE}}
\label{proof:twoODE}
\begin{lemma*} 
Consider the following system of ODEs.
    \begin{align}
        \epsilon_1'(t) &= (A_{1,1}t+B_{1,1}) \epsilon_1(t) + (A_{1,2}t+B_{1,2}) \epsilon_2(t),\\
        \epsilon_2'(t) &= (A_{2,1}t+B_{2,1}) \epsilon_1(t) + (A_{2,2}t+B_{2,2}) \epsilon_2(t), 
    \end{align}
Then we have 
\begin{align*}
    \epsilon_1''(t) = P_1(t) \epsilon_1'(t) + P_2(t) \epsilon_1(t),
\end{align*}
where
\begin{align*}
    P_1(t) &= A_{1,1}t+B_{1,1} + (A_{1,2}t+B_{1,2})^{-1}\big( A_{1,2}+(A_{1,2}t+B_{1,2})(A_{2,2}t+B_{2,2}) \big), \\
    P_2(t) &= \big(A_{1,1} + (A_{1,2}t+B_{1,2})(A_{2,1}t+B_{2,1}) \big), \\
    &- (A_{1,2}t+B_{1,2})^{-1}\big( A_{1,2}+(A_{1,2}t+B_{1,2})(A_{2,2}t+B_{2,2}) \big)  (A_{1,1}t+B_{1,1}).
\end{align*}
\end{lemma*}
\begin{proof}
First, from \eqref{eq:e1p}, we have 
\begin{align}
    \epsilon_2(t) =  (A_{1,2}t+B_{1,2})^{-1} \big( \epsilon_1'(t) - (A_{1,1}t+B_{1,1}) \epsilon_1(t) \big).  \label{eq:eps2}
\end{align}
    By taking the derivative to \eqref{eq:e1p}, we have 
\begin{align*}
    &\epsilon_1''(t) = (A_{1,1}t+B_{1,1}) \epsilon_1'(t) + A_{1,1}\epsilon_1(t)+ (A_{1,2}t+B_{1,2}) \epsilon_2'(t) + A_{1,2}\epsilon_2(t) \\
    &= (A_{1,1}t+B_{1,1}) \epsilon_1'(t) + A_{1,1}\epsilon_1(t)+ (A_{1,2}t+B_{1,2}) \big((A_{2,1}t+B_{2,1}) \epsilon_1(t) + (A_{2,2}t+B_{2,2}) \epsilon_2(t) \big) + A_{1,2}\epsilon_2(t) \\
    &= (A_{1,1}t+B_{1,1}) \epsilon_1'(t)+ \big(A_{1,1} + (A_{1,2}t+B_{1,2})(A_{2,1}t+B_{2,1}) \big) \epsilon_1(t) + \big( A_{1,2}+(A_{1,2}t+B_{1,2})(A_{2,2}t+B_{2,2}) \big)\epsilon_2(t) \\
    &= (A_{1,1}t+B_{1,1}) \epsilon_1'(t)+ \big(A_{1,1} + (A_{1,2}t+B_{1,2})(A_{2,1}t+B_{2,1}) \big) \epsilon_1(t) \\
    & + \big( A_{1,2}+(A_{1,2}t+B_{1,2})(A_{2,2}t+B_{2,2}) \big)(A_{1,2}t+B_{1,2})^{-1} \big( \epsilon_1'(t) - (A_{1,1}t+B_{1,1}) \epsilon_1(t) \big),
\end{align*}
where we use \eqref{eq:eps2} for the last equality. Hence we achieve the result.
\end{proof}

\subsection{Proof of \cref{lemma:Ntau}}
\label{proof:Ntau}
\begin{lemma*}
Let
\begin{align*}
    N(\tau) &= -\frac{1}{2}A_{1,2}^{-1}\big(A_{1,2}+(A_{1,2}\tau +B_{1,2})(A_{2,2}\tau) \big)  \\
    & -\frac{1}{2} A_{1,2}^{-2} (A_{1,2}\tau + B_{1,2})( 2(A_{1,2}+A_{2,2})\tau + B_{1,2}A_{2,2} )  \\
    &+\frac{1}{2}A_{1,2}^{-2}(A_{1,1}\tau +B_{1,1})(A_{1,2}\tau + B_{1,2})\big(A_{1,2}+(A_{1,2}\tau +B_{1,2})(A_{2,2}\tau) \big)  \\
    &+\frac{1}{4}A_{1,2}^{-2} \big(A_{1,2}+(A_{1,2}\tau +B_{1,2})(A_{2,2}\tau) \big)^2 \\
    &-A_{1,2}^{-2} (A_{1,2}\tau + B_{1,2})(A_{1,1}\tau + B_{1,1})  \big(A_{1,2}+(A_{1,2}\tau +B_{1,2})(A_{2,2}\tau) \big). \nonumber
\end{align*}
Then we have 
\begin{align*}
N(\tau) =  N_2(\tau)(\tau + A_{1,2}^{-1} B_{1,2})^2 + N_1(\tau) (\tau + A_{1,2}^{-1} B_{1,2}) + N_0(\tau),
\end{align*}
where
\begin{align*}
    N_2(\tau) &=(\frac{1}{4}A_{2,2}-\frac{1}{2}A_{1,1}A_{2,2} )\tau^2 -(\frac{1}{2}B_{1,1}A_{2,2}+1)\tau -\frac{1}{2}A_{1,1},   \\
    N_1(\tau) &=  \frac{1}{2}A_{1,2}^{-1}B_{1,2}( A_{1,1} - A_{2,2} ) -\frac{1}{2}B_{1,1},  \\
    N_0(\tau) &= -\frac{1}{4}.
\end{align*}
\end{lemma*}

\begin{proof}
    An expansion of $N(\tau)$ gives 
\begin{align*}
N(\tau) = M_2(\tau) (\tau + A_{1,2}^{-1} B_{1,2})^2 + M_1(\tau) (\tau + A_{1,2}^{-1} B_{1,2}) + M_0(\tau),
\end{align*}
where 
\begin{align*}
    M_2(\tau) &= - \tau   +\frac{1}{2}(A_{1,1}\tau +B_{1,1})A_{2,2}\tau + \frac{1}{4}A_{2,2} \tau^2  - (A_{1,1}\tau+B_{1,1})A_{2,2}\tau  ,  \\
    M_1(\tau) &= -\frac{1}{2} A_{2,2} \tau  - \frac{1}{2}A_{1,2}^{-1}B_{1,2}A_{2,2} + \frac{1}{2} (A_{1,1}\tau +B_{1,1}) +\frac{1}{2}A_{2,2}\tau - (A_{1,1}\tau+B_{1,1}),  \\
    M_0(\tau) &= -\frac{1}{2} + \frac{1}{4} = - \frac{1}{4}.  
\end{align*}
Since 
\begin{align*}
    M_2(\tau) &= \frac{1}{4} A_{2,2} \tau^2 -\frac{1}{2}(A_{1,1}\tau+B_{1,1})A_{2,2}\tau  - \tau =  (\frac{1}{4}A_{2,2}-\frac{1}{2}A_{1,1}A_{2,2} )\tau^2 -(\frac{1}{2}B_{1,1}A_{2,2}+1)\tau,  \\
    M_1(\tau) &=  - \frac{1}{2}A_{1,1}  \tau - \frac{1}{2}A_{1,2}^{-1}B_{1,2}A_{2,2}  -\frac{1}{2}B_{1,1}  \\
    &=  - \frac{1}{2} A_{1,1}  (\tau + A_{1,2}^{-1} B_{1,2}) +  \frac{1}{2}A_{1,2}^{-1}B_{1,2}( A_{1,1} - A_{2,2} ) -\frac{1}{2}B_{1,1},
\end{align*}
We achieve the desired result. 
\end{proof}

\bibliographystyle{unsrt}  
\bibliography{references}  

\end{document}